\documentclass[final,3p,times,authoryear]{elsarticle}
\usepackage{amssymb}
\usepackage{amsthm}
\usepackage{amsmath}
\usepackage{booktabs}
\usepackage{multirow}
\usepackage{textcomp}
\usepackage{subcaption} 
\usepackage{siunitx}
\usepackage{epstopdf}
\usepackage[table,xcdraw]{xcolor}
\usepackage[colorlinks=true, linkcolor=blue]{hyperref}
\usepackage{url}
\usepackage[linesnumbered,ruled,vlined]{algorithm2e}
\usepackage{subcaption}
\usepackage{float}
\usepackage[authoryear]{natbib}
\usepackage{threeparttable}
\usepackage{xcolor} 
\usepackage{hyperref}
\usepackage[table]{xcolor}
\newcommand{\rev}[1]{\textcolor{yishen}{#1}} 
\newcommand{\revv}[1]{\textcolor{green}{#1}} 
\newcommand{\revvv}[1]{\textcolor{ershen}{#1}} 
\definecolor{ershen}{RGB}{0,0,0}
\definecolor{green}{RGB}{0,0,0}
\definecolor{yishen}{RGB}{0,0,0}
\setcitestyle{authoryear, round}
\newcommand{\tablesize}{\fontsize{8}{10.5}\selectfont}

\journal{}

\begin{document}

\begin{frontmatter}


\title{Learning to Rank Critical Road Segments via Heterogeneous Graphs with Origin-Destination Flow Integration}


\author[th]{Ming Xu\corref{cor1}}
\ead{xum.2016@tsinghua.org.cn}
\cortext[cor1]{Corresponding author}
\affiliation[th]{organization={School of Software, Liaoning Technical University},
            postcode={125105},
            city={Huludao},
            country={China}}

\author[th]{Jinrong Xiang}
\ead{xiangjinrong117@gmail.com}

\author[th]{Zilong Xie}
\ead{zilong6037@gmail.com}

\author[pku]{Xiangfu Meng}
\ead{marxi@126.com}
\affiliation[pku]{organization={Faculty of Electronic and Information Engineering, Liaoning Technical University},
            postcode={125105},
            city={Huludao},
            country={China}}%

\footnote{Our code and data are available at : \url{https://github.com/iCityLab/HetGL2R.}}

\begin{abstract}
\rev{Existing learning-to-rank methods for road networks often fail to incorporate origin–destination (OD) flows and route information, limiting their ability to model long-range spatial dependencies. To address this gap, we propose HetGL2R, a heterogeneous graph learning framework for ranking road-segment importance. HetGL2R builds a tripartite graph that unifies OD flows, routes, and network topology, and further introduces attribute-guided graphs that elevate node attributes into explicit nodes to model functional similarity. A heterogeneous joint random walk algorithm (HetGWalk) jointly samples both graph types to generate context-rich node sequences. These sequences are encoded using a Transformer to learn embeddings that capture long-range structural dependencies induced by OD flows and route configurations, as well as functional associations derived from attribute similarity. Finally, a listwise ranking strategy with a KL-divergence loss evaluates and ranks segment importance. Experiments on three SUMO-generated simulated networks of different scales show that, against state-of-the-art methods, HetGL2R achieves average improvements of approximately 7.52\%, 4.40\% and 3.57\% in ranking performance.}
\end{abstract}




\end{frontmatter}

\section{Introduction}
\label{sec:Introduction}
\rev{Efficient and resilient road networks are essential for ensuring smooth urban mobility and public safety. When a single road segment becomes congested or blocked, the resulting disruption often propagates along multiple routes, leading to large-scale delays or even citywide paralysis. Therefore, identifying critical road segments—those whose failure would significantly degrade overall network performance—is of great importance for traffic management and infrastructure planning \citep{rf1}.}

\begin{figure}[htbp]
    \centering
    \includegraphics[width=0.8\linewidth]{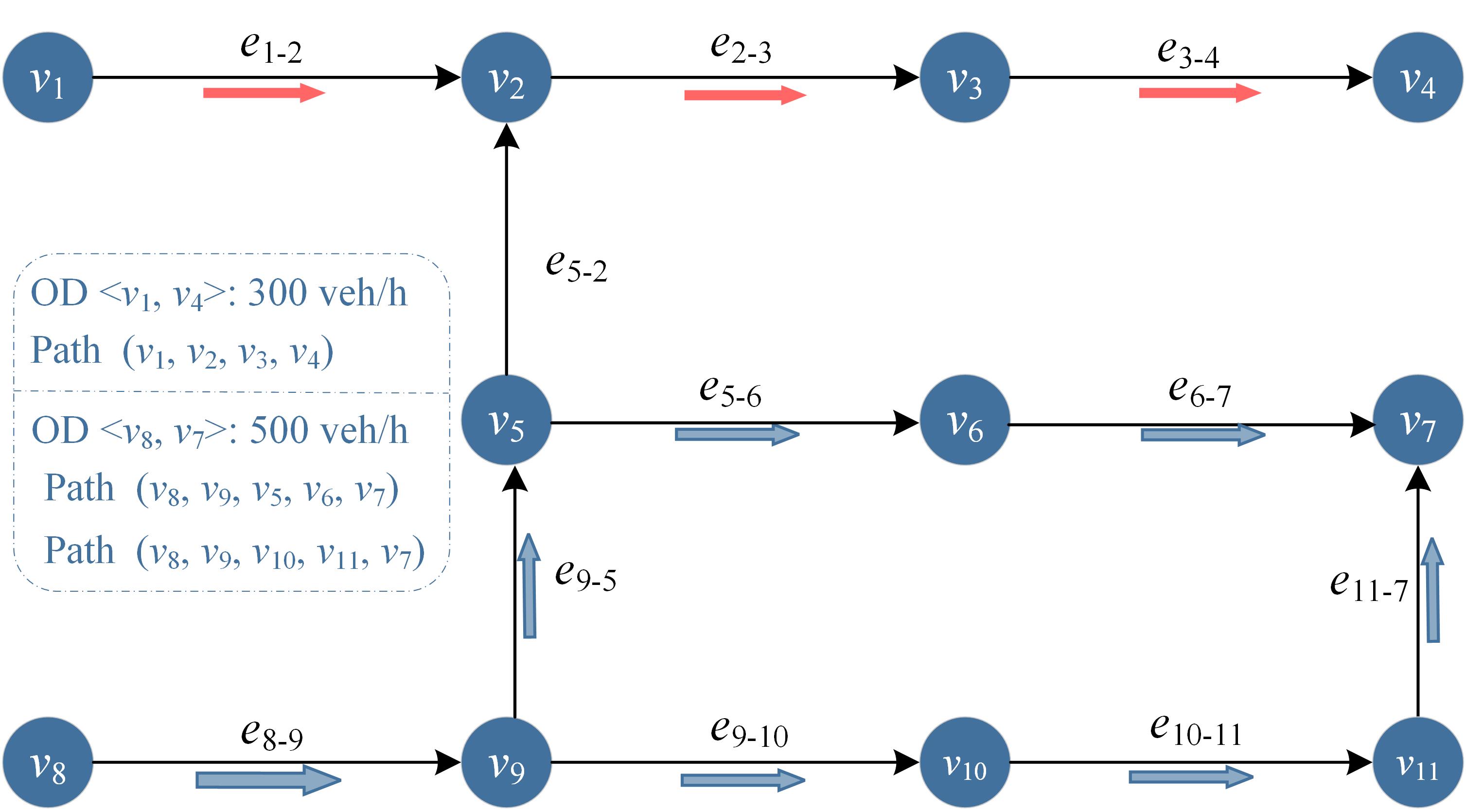}
    \caption{A small simulated road network used for illustration. The network consists of 11 nodes ($v$) and 11 directed edges ($e$), with two origin–destination (OD) pairs: $<v_1$, $v_4>$ with 300 vehicles traveling along a single path, and $<v_8$, $v_7>$ with 500 vehicles distributed over two alternative paths. Red and blue arrows indicate the routing paths associated with the two OD pairs, respectively.}
    \label{fig:g7}
\end{figure}

\rev{Prior studies on road-network criticality were primarily based on complex network theory, in which centrality measures (e.g., early betweenness centrality \citep{betweenness} and PageRank \citep{rf3}, and more recent methods such as IDME \citep{IDME}, EPC \citep{EPC}, OVED-Rank \citep{rf44}, LSS \citep{rf42}, and GLC \citep{rf41}) were used to quantify the structural importance of nodes. These approaches are intuitive and easy to interpret but fail to incorporate the rich attribute features and dynamic traffic behaviors associated with each road segment. In reality, a segment’s criticality depends on multiple factors such as traffic volume, number of lanes, and functional hierarchy, many of which are neglected in purely topological metrics. Some studies have attempted to combine centrality measures with attribute features (e.g., study \citep{rf3}), but most rely on manually tuned parameters, limiting their scalability and generalization to large real-world networks.}

\rev{In recent years, graph neural network (GNN)–based learning methods \citep{rf4,rf8} have attracted increasing attention in the field of node importance assessment. These models integrate node attributes with local topological information through message-passing mechanisms to learn effective node embeddings. However, due to the limited receptive field of GNNs, they struggle to capture the long-range path dependencies commonly present in traffic flows, which limits their performance in critical road segment ranking tasks.}

\rev{In transportation networks, the global influence of a road segment is largely shaped by origin–destination (OD) flows, which quantify the travel demand between each OD pair. The OD flows determine which paths are utilized and how traffic load is distributed, thereby shaping each segment’s functional role and criticality \citep{od1,od2,od3}. For instance, if a segment lies on the shortest paths of multiple high-demand OD pairs, any incident or restriction there can induce congestion that propagates upstream along the affected OD paths, leading to severe network-wide delays. Conversely, a peripheral segment carrying a similar traffic volume but serving only a few OD flows would have a far more limited systemic impact if disrupted.}

\rev{As illustrated in Figure \ref{fig:g7}, a simulated road network contains two OD flows, $< v_1$, $v_4>$ and $<v_8$, $v_7>$. When an incident occurs at node $v_{11}$, congestion propagates upstream along the path of OD flow $<v_8$, $v_7>$ to nodes $v_{10}$ and $v_9$. This forces some travelers to reroute, thereby increasing the congestion risk on distant segments $v_5$ and $v_6$. In contrast, segments $v_2$ and $v_3$ remain largely unaffected as they do not carry the affected OD flows. This example highlights that the influence among road segments is fundamentally governed by the OD flows they serve and the paths they form. Conceptually, spatial dependencies in a road network stem from functional interactions among OD flows, paths, and segments, rather than from physical connectivity. Consequently, the OD-dominated functional relationships are pivotal for evaluating node criticality.}

\rev{Existing graph learning approaches face two fundamental limitations in this task. First, they primarily learn structural representations from the physical topology of road networks while overlooking the functional structures shaped by OD flows. Second, their local neighborhood aggregation mechanism is constrained by network depth, resulting in limited receptive fields. As a result of these limitations, they struggle to capture long-range spatial dependencies and consequently fail to accurately model the impact patterns of local disruptions (e.g., an incident at node $v_{11}$)—where the scope of impact is dictated by the functional roles of segments within OD flows and paths, rather than by physical proximity. Although CRRank \citep{rf1} incorporates OD information to evaluate the functional importance of road segments, its reliance on manually designed aggregation rules and similarity metrics constrains its expressiveness and effectiveness, as it cannot learn complex interaction patterns in an end-to-end manner.}

\rev{To overcome these limitations, this paper proposes HetGL2R, a heterogeneous graph learning framework that integrates OD flow information for road-segment ranking. The key contributions of this study can be summarized as follows:}

\begin{itemize}
\setlength{\itemsep}{0pt}
    \item \rev{\textbf{Methodological Innovation}: HetGL2R constructs a tripartite graph (connecting OD pairs, paths, and road segments) to integrate the functional structure of the road network with its physical structure. Simultaneously, it introduces three attribute-node bipartite graphs—one each for OD pairs, paths, and segments—to explicitly model the attribute-based functional similarities. A joint random walk mechanism operates on these graphs collectively to sample context-rich node sequences, which are then processed by a Transformer encoder to learn the long-range dependencies and functional roles of segments for accurate criticality ranking.}

    \item \rev{\textbf{Theoretical Insight}: Our core insight is that spatial dependencies in traffic networks are governed by functional structure rather than physical topology. To capture this structure, we introduce a joint random walk that integrates two complementary perspectives: functional role relationships (OD-path-segment) and attribute-based functional similarities. Consequently, the model learns the underlying mechanisms of spatial dependency, thereby enabling accurate assessment of nodes' spatial influence.}

    \item \rev{\textbf{Practical Implication}: Extensive evaluation on three simulated traffic network datasets of varying scales demonstrates that our proposed framework consistently outperforms existing baseline methods. The framework can serve as a foundation for applications such as transportation resilience analysis, emergency route planning, and infrastructure maintenance prioritization.}
\end{itemize}

\section{Related Works}
In this section, we discuss three research areas that are closely related to this paper.
\subsection{Evaluation of node importance in networks}
Existing research on identifying critical nodes in networks can be broadly categorized into two approaches: centrality-based methods and machine learning-based methods. The limitation of centrality-based methods lies in their inability to fully leverage the multidimensional features of nodes. Although study \citep{rf3} attempts to improve critical node identification in road networks by integrating node centrality with attribute features, its performance heavily depends on manually set topology-feature trade-off parameters. This leads to a lack of generalizability in quantifying the contributions of topology and features, limiting its applicability in large-scale real-world road networks. To address these limitations, LSS \citep{rf42} estimates node importance based on local information, eliminating the need for global knowledge and complex parameter tuning. Study \citep{rf41} proposed a dual-perspective approach that identifies influential nodes by considering inter-cluster connectivity and local neighborhood importance. However, similar to traditional centrality-based methods, these approaches still fail to incorporate node attribute features or dynamic network behaviors, limiting their applicability in more complex real-world scenarios such as road networks. In recent years, researchers have gradually shifted towards machine learning-based methods, which can be divided into two main technical approaches: reinforcement learning frameworks (e.g., FINDER proposed in study \citep{rf4}, which utilizes GraphSAGE \citep{rf5} to learn node importance; study \citep{rf6}, which combines attribute-aware graph attention networks with double DQN to optimize critical node identification in sparse graphs) and graph representation learning (e.g., TraNode2vec from study \citep{rf7}, which assesses node importance by integrating topological embedding clustering and traffic flow; study \citep{rf8}, which introduces the MGL2Rank framework that employs multi-graph fusion techniques to jointly model topology and road segment attributes). However, when applying these methods to road network node importance evaluation, they fail to incorporate OD flow and path selection. As we analyzed in the introduction, these two types of information are crucial for assessing node importance in road networks. Therefore, there remains significant room for improvement in the performance of these methods.

\subsection{Learning to rank}
Learning to rank (LTR) originated in the field of information retrieval and is a technique for ordering documents, items, or other entities based on their relevance. In recent years, it has been widely applied in various domains, including recommender systems \citep{rf9}, sentiment analysis \citep{rf10}, and object detection \citep{rf11}. Depending on the characteristics of the input and output spaces, LTR methods can be broadly categorized into three types: pointwise, pairwise, and listwise approaches. The pointwise approach \citep{rf12} treats each item independently, formulating the ranking task as a classification or regression problem. However, this method neglects the relative relationships among items. In contrast, the pairwise approach \citep{rf13} transforms the ranking task into a binary classification problem by comparing item pairs to determine their relative order. Prior research \citep{rf14} has introduced graph neural networks into pairwise ranking to approximate the importance of nodes within a network. Although the pairwise method offers improved accuracy over the pointwise method, it still fails to fully capture the global context and relative importance of items within the entire set, limiting its overall performance. The listwise approach \citep{rf15} treats ranking as a holistic optimization problem and has become a focal point of research due to its modeling and computational complexity. For example, one study \citep{rf16} analyzes how list length and loss function selection affect algorithm stability through generalization error bounds, while another \citep{rf17} creatively reformulates ranking as a multi-task regression problem, overcoming the limitations of traditional linear models. With the rise of deep learning, researchers \citep{rf18,rf19,rf20} have incorporated attention mechanisms and recurrent neural networks into listwise modeling, significantly enhancing performance in scenarios involving complex interactions. On the application side, DiAL \citep{rf21} is the first to integrate the $\alpha$-NDCG diversity metric into listwise loss functions. Further advancements include leveraging large language models to generate reference answer sequences for listwise evaluation  \citep{rf22,rf45}, as well as developing differentiable approximations to ranking indicators that closely match true loss functions with limited data \citep{rf23}. Recent work \citep{rf24} also proposes a regression alignment strategy to address the multi-objective trade-off between score calibration and ranking optimization. Building upon the core principles of listwise ranking, this paper proposes a heterogeneous graph learning framework tailored for node ranking in transportation networks.

\subsection{Heterogeneous graph learning}
Heterogeneous graphs \citep{rf25} model multiple types of nodes and edges, integrating information from heterogeneous data sources, thereby demonstrating a stronger representation ability for complex systems. Representation learning on heterogeneous graphs aims to obtain meaningful node embeddings to support various downstream tasks, such as node ranking and link prediction \citep{rf26,rf27}. Many studies have focused on designing graph neural network models suitable for heterogeneous graphs. For example, most of these networks \citep{rf28,rf29,rf30} utilize meta-paths and type-aware aggregation strategies to capture the heterogeneity of the graph effectively, that is, the difference in the type of nodes and edges. HetGNN \citep{rf31} combines random walk strategies with bidirectional LSTM structures, effectively integrating heterogeneous node attributes and their neighboring information. HetSANN \citep{rf32} directly encodes the information of different types of nodes and edges through a type-aware attention mechanism, enhancing the model's ability to represent the heterogeneous graph structure. These methods are distinctive in their modeling strategies, but most still rely on message-passing mechanisms, which do not fully exploit the potential structural relationships between nodes in heterogeneous graphs. To address this, M2CHGNN \citep{rf43} employs a dual-view encoding strategy that integrates attention mechanisms with contrastive learning to generate high-quality node representations for heterogeneous graphs. UI-HEHo \citep{rf47} effectively learns high-order node embeddings via multi-head attention mechanisms and multi-task shared features. SchemaWalk \citep{rf34} proposes an unbiased edge-type exploration strategy, providing finer-grained and more generalizable methods for embedding heterogeneous networks. JUST \citep{rf35} introduces the "Jump-Stay" strategy for adaptive node sampling, further improving the representation capability. HMSG \citep{rf36} constructs a meta-path-based multi-relational graph neural network, and employs contrastive learning together with Deep Graph Infomax (DGI) \citep{rf37} to jointly learn high-quality node representations. HPSRW \citep{rf38} combines higher-order Markov chains and the Spacey Walk method to effectively mitigate the stability issues in meta-path sampling. \rev{While heterogeneous graph learning has advanced in various domains, existing general-purpose models lack specialized designs for the unique characteristics of transportation networks. Even when applied to our constructed graph that explicitly incorporates OD flows and route information, these models still fail to effectively leverage its structural properties, leading to limited capability in capturing the spatial dependencies inherent in road networks.} To the best of our knowledge, CRRank \citep{rf1} is the first work to apply heterogeneous graphs to transportation network analysis, achieving node ranking by integrating OD flow and path information. However, this method relies on manually defined computational rules and lacks end-to-end learnability, leaving room for substantial improvements in model performance.

\section{\rev{Preliminary}}

\subsection{\rev{Definitions and problem statement}}
\label{sec:3.1}

Before presenting our method, we first provide the definitions relevant to this work.

\textbf{Definition 1 }(\textit{Road Network}). A road network is modeled as a directed graph $G(V^l,\rev{E^l})$, where $V^l$ denotes the set of nodes and \rev{$E^l$} denotes the set of directed edges. In this paper, each node $v^l_i \in V^l$ represents a road segment, and a directed edge $e_{ij}^l \in V^l$ indicates a directed connection from $v^l_i$ to $v^l_j$. Each node $v^l_i$ is associated with an initial attribute vector describing its physical or traffic-related properties.

\textbf{Definition 2 }(\textit{OD Flow}). An OD flow is defined as a triple \rev{$〈v_i^l,v_j^l,f〉$, where $v_i^l, v_j^l\in V^l$ denote the origin and destination nodes, respectively, and $f$ denotes the number of vehicles from $v_i^l$ to $v_j^l$ over a predefined observation period. For instance, the road network in Figure \ref{fig:g7} contains two OD flows: $<v_1, v_4, 300>$ and $<v_8, v_7, 500>$.}

\textbf{Definition 3 }(\textit{Path}). A path $\rho=(v^l_1, v^l_2,...,v^l_{|\rho|} )$ is defined as an ordered sequence of road-segment nodes in the road network, representing the segments traversed by trips associated with a specific origin–destination (OD) flow. Here, $|\rho|$ denotes the number of road segments in path $\rho$. In practice, a single OD flow may be associated with multiple feasible paths. For example, the OD pair $<v_8, v_7,500>$ corresponds to two paths: $(v_8, v_9, v_5, v_6, v_7)$ and $(v_8, v_9, v_{10}, v_{11}, v_7)$.

\textbf{Definition 4 }(\textit{Trip Graph}). A trip graph is a heterogeneous graph $TG(V, A, M)$, where $V=V^o\cup V^p\cup V^l$ consists of OD nodes $V^o$, path nodes $V^p$, and road-segment nodes $V^l$. Each node type is associated with an attribute matrix, denoted by $A^o$, $A^p$, and $A^l$. The adjacency structure $M$ comprises three components: (1) OD–Path association matrix $M^{o,p} \in \mathbb{R}^{|V^o| \times |V^p|}$, where each entry $m^{o,p}\in [0,1]$ represents the traffic proportion of an OD pair assigned to a specific path; (2) Path–Segment incidence matrix $M^{p,l} \in \left\{0,1\right\}^{|V^p| \times |V^l|}$, where $m^{p,l}=\mathbb{I}(v^l\in \rho)$ indicates whether road segment $v^l$ is contained in path $\rho$. (3) Segment–Segment connectivity matrix $M^{l,l} \in \mathbb{R}^{|V^l| \times |V^l|}$, whose elements are defined as

\begin{equation}
m_{i,j}^{l,l}
=\sum_{\substack{\rho\in(\mathcal{P}_i\cap\mathcal{P}_j)\\[2pt]
v_i^{l}\prec_\rho v_j^{l}}}
e^{-c_{i,j}^{\rho}}
\label{eq:e1}
\end{equation}
where $\mathcal{P}_i$ and $\mathcal{P}_j$ denote the sets of paths containing $v_i^l$ and $v_j^l$, respectively. The notation $v_i^l \prec_\rho v_j^l$ indicates that segment $v_i^l$ lies upstream of $v_j^l$ along path $\rho$, $c_{i,j}^\rho$ is the number of intermediate segments between them on $\rho$. If no such path exists, $m_{i,j}^{l,l}=0$. $e$ is a propagation decay factor. This formulation models the strength of influence propagation between segments along OD-induced paths.

\rev{While the Trip Graph (TG) effectively integrates the macro-level functional organization and physical structure of the road network, it is less effective at capturing local, attribute-based semantic similarities among nodes of the same type. For instance, the functional similarity between OD pairs that both connect residential and commercial areas, or between road segments sharing comparable attributes like the number of lanes or traffic capacity, is crucial for learning comprehensive and robust segment representations. To address this limitation, we introduce the Attribute-Guided Graph (AG) as a complementary semantic structure to the TG. Its core innovation lies in explicitly representing node attributes as nodes, thereby constructing a heterogeneous bipartite graph that directly models these attribute-based functional relationships.}

\rev{\textbf{Definition 5 }(\textit{Attribute-Guided Graph}). For each node type $\phi \in \left\{ o, p, l \right\}$, where $o$, $p$ and $l$ denote OD pairs, paths, and road segments respectively, an Attribute-Guided Graph is defined as a weighted bipartite graph $AG_{\phi} (V^{\phi} \cup U^{\phi},A^{\phi})$. Here, $V^{\phi}$ denotes the set of nodes of type $\phi$, $U^{\phi}$ denotes the set of attribute nodes associated with type $\phi$, and $A^{\phi}$ denotes the set of edges connecting $v_i^{\phi} \in V^{\phi}$ and $u_k^{\phi} \in U^{\phi}$. Each edge $(v_i^{\phi},u_k^{\phi}) \in A^\phi$ is assigned a weight $a_{ik}^{\phi}$, which reflects the similarity between node $v_i^{\phi}$ and attribute $u_k^{\phi}$. Special cases include $AG_o (V^o \cup U^o,A^o)$ for OD pairs, $AG_p (V^p \cup U^p,A^p)$ for paths, and $AG_l (V^l \cup U^l,A^l)$ for road segments.}

\textbf{Definition 6 }(\textit{Learning to Rank Road Segments}). Let each road segment $v_i^l\in V^l$ be represented by a feature vector $h_i$. Given a list of $K$ road segments with representations $H=\left[ h_1,h_2, \cdots,h_K\right]$, and their corresponding ground-truth importance ordering $Y=\left[ y_1,y_2, \cdots,y_K\right]$, where $y_i$ denotes the rank position of segment $v_i^l$ and higher-ranked segments are considered more critical, the learning-to-rank task aims to learn a parametric mapping $F_\theta:\mathbb{R}^{K \times d}\rightarrow Y^K$, where $Y^K$ denotes the set of all possible permutations of $K$ elements. The output $\hat{Y}=F_\theta(H)$ represents the predicted ranking of the input road segments. The parameters $\theta$ are learned by minimizing a listwise ranking loss that measures the discrepancy between the predicted ranking $\hat{Y}$ and the ground-truth ordering $Y$.

\begin{table}[t]
\centering
\small
\caption{\rev{Notation and their definitions.}}
\label{notation}
\setlength{\tabcolsep}{15pt}
\renewcommand{\arraystretch}{1.2}
\begin{tabular}{p{0.15\textwidth} p{0.72\textwidth}}
\toprule \hline
\rev{\textbf{Symbol}} & \rev{\textbf{Meaning}} \\
\hline
$G, \,TG, \,AG$ & road network graph, trip graph, attribute-guide graph \\
$o, \,p, \, l$ & OD type, path type, segment type \\
$\phi$ & node type, $\phi \in \{o, \,p, \, l\}$ \\
$V^\phi, \,v^\phi$ & set of nodes of type $\phi$, a node of type $\phi$ \\
$A^\phi$ & attribute matrix for nodes of type $\phi$ \\
$U^\phi, \,u^\phi$ & set of attribute nodes of type $\phi$, an attribute node of type $\phi$ \\
$V, \,A, \,U$ & $V^{o}\cup V^{p}\cup V^{l}$, $A^{o}\cup A^{p}\cup A^{l}$, $U^{o}\cup U^{p}\cup U^{l}$ \\
$\rho$ & a path \\
\revv{$\mathcal{P}_i$} & \revv{set of paths containing segment $i$} \\
\revv{$a_{ij}^\phi, \,\bar{a}_{ij}^\phi$} & \revv{$j$-th attribute value of node $i$ of type $\phi$ and its normalized form} \\
$M^{o,p}$ & OD–Path association matrix \\
$M^{p,l}, \,\revv{\bar{M}^{p,l}, \, \bar{M}'^{p,l}}$ & path–segment incidence matrix \revv{and its row- and column-normalized forms} \\
$M^{l,l},\revv{\bar{M}^{l,l}}$ & segment–segment connectivity matrix \revv{and its row-normalized form} \\
\revv{$H$} & \revv{list of road segment representations used for ranking} \\
\revv{$K$} & \revv{number of segments in $H$} \\
\revv{$h_i$} & \revv{representation of segment $i$} \\
$num, \,len$ & number of generated sequences per OD, length of each sequence \\
$\alpha$ & probability of choosing TG-based transitions in the walk \\
$\beta$ & trade-off parameter between depth-first and breadth-first transitions on the TG \\
\revv{$\varepsilon$} & \revv{trade-off parameter between segment–segment and path–segment transitions on the TG} \\
$Y, \,\hat{Y}$ & ground-truth and predicted rank position vectors of a segment list \\
$y_i, \, \hat{y}_i$ & ground-truth and predicted rank positions of segment $i$ \\
\hline \bottomrule
\end{tabular}
\end{table}

\rev{In summary, the above definitions provide the structural and semantic foundations for modeling road networks from a functional perspective. By jointly considering OD flows, paths, and attribute-based similarities, they enable representation learning methods to go beyond purely topological dependencies and capture the functional interactions that govern long-range spatial influence in traffic networks. Table \ref{notation} provides a summary of key notation and their definitions.}

\subsection{\rev{Research objectives}}
\revv{Building upon the problem definition in Section \ref{sec:3.1}, this study is guided by the following research objectives:}

\rev{\textbf{(RO1)} To design a joint heterogeneous graph random walk  mechanism that generates context-rich sequences by \revv{jointly sampling OD-induced} functional structures and attribute-based similarities, \revv{enabling the model to encode both functional interactions and semantic consistency among road segments.}}

\rev{\textbf{(RO2)} To formulate a learning-to-rank framework that leverages the functional structure of road networks\revv{-induced by OD flows and path interactions-}to achieve accurate prioritization of the critical \revv{road segments.}}

\revv{\textbf{(RO3)} To examine the contribution of functional-structure modeling to road-network ranking performance through comparisons with topology-driven baselines, thereby clarifying the role of OD-induced functional dependencies beyond purely structural cues.}

\revv{These objectives directly inform the design of the proposed HetGL2R framework and the subsequent experimental evaluation presented in the following sections.}

\section{Methodology}

\begin{figure}[htbp]
    \centering
    \includegraphics[width=1\linewidth]{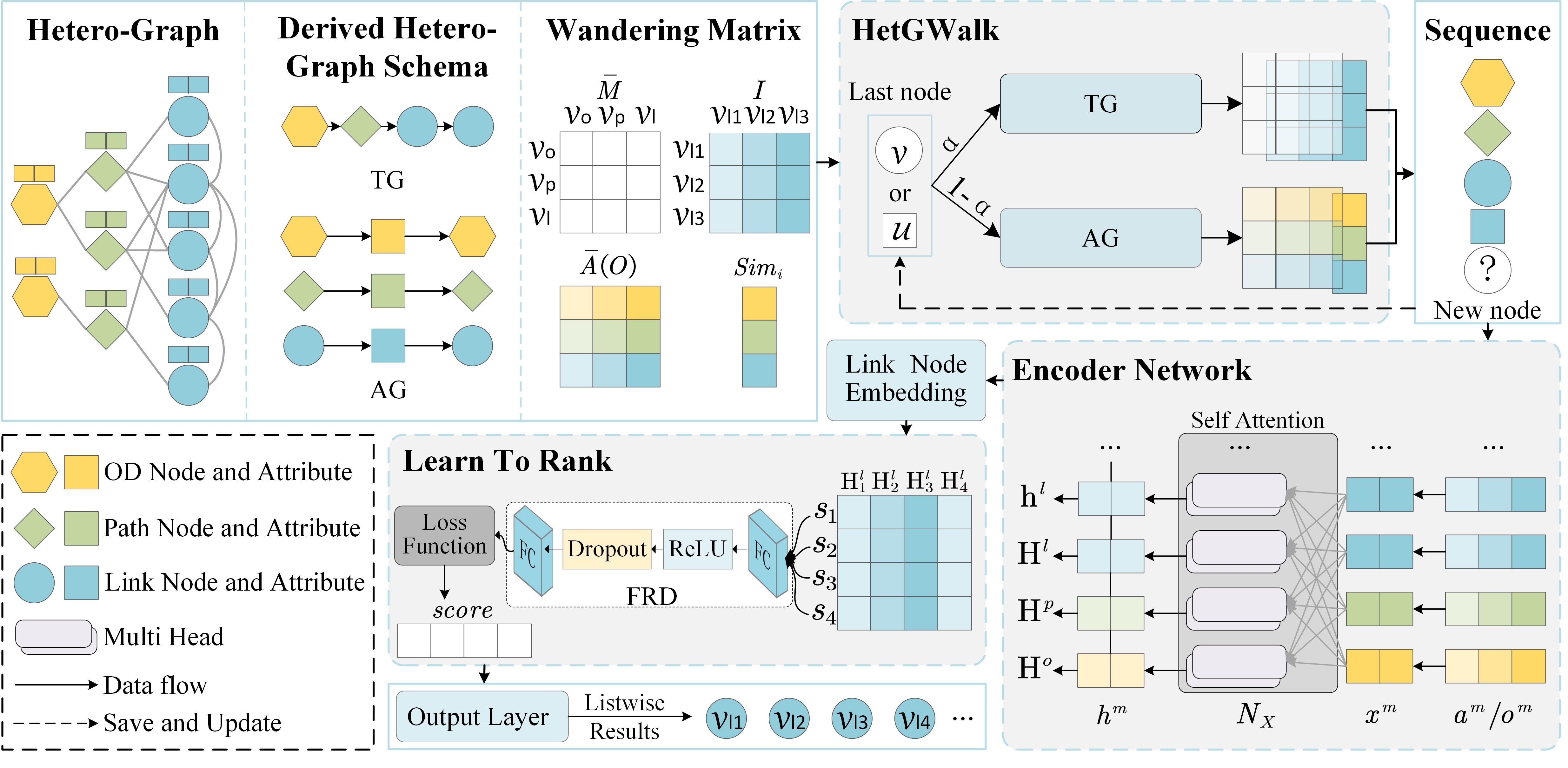}
    \caption{\revv{Framework of the proposed HetGL2R method. The framework consists of four main stages: heterogeneous graph construction, sequence generation via HetGWalk, feature encoding, and learning to rank. Yellow, green, and blue nodes represent origin–destination (OD) pairs, paths, and road segments, respectively. Solid arrows indicate the main data flow, while dashed arrows denote the saving and updating processes during model training. Specifically, HetGWalk samples from the trip graph and attribute-guided graphs to generate heterogeneous sequences, which are encoded by a Transformer-based encoder and subsequently used for listwise ranking.}}
    \label{fig:g1}
\end{figure}

\rev{This paper proposes HetGL2R, a deep learning framework (Figure \ref{fig:g1}) based on random walks on heterogeneous graphs, to improve traffic network ranking by integrating their functional structures with physical topology. The framework first extracts information from vehicle trajectory data to construct an OD-path-segment tripartite graph, which integrates the functional and physical structures of the road network into a unified representation. Simultaneously, a set of three attribute-guided bipartite graphs—one each for OD pairs, paths, and segments—are constructed to model functional similarities among nodes. Subsequently, the heterogeneous graph joint random walk algorithm, HetGWalk, is applied to these graphs to sample context-rich node sequences. These sequences are encoded using a Transformer encoder to learn road segment embeddings that capture long-range dependencies. Finally, a ranking learning module comprehensively evaluates and ranks the importance of road segments.}

\subsection{HetGWalk algorithm}
\subsubsection{\revv{Joint Random Walk Procedure}}
\begin{figure}[ht]
    \centering
    \includegraphics[width=1\linewidth]{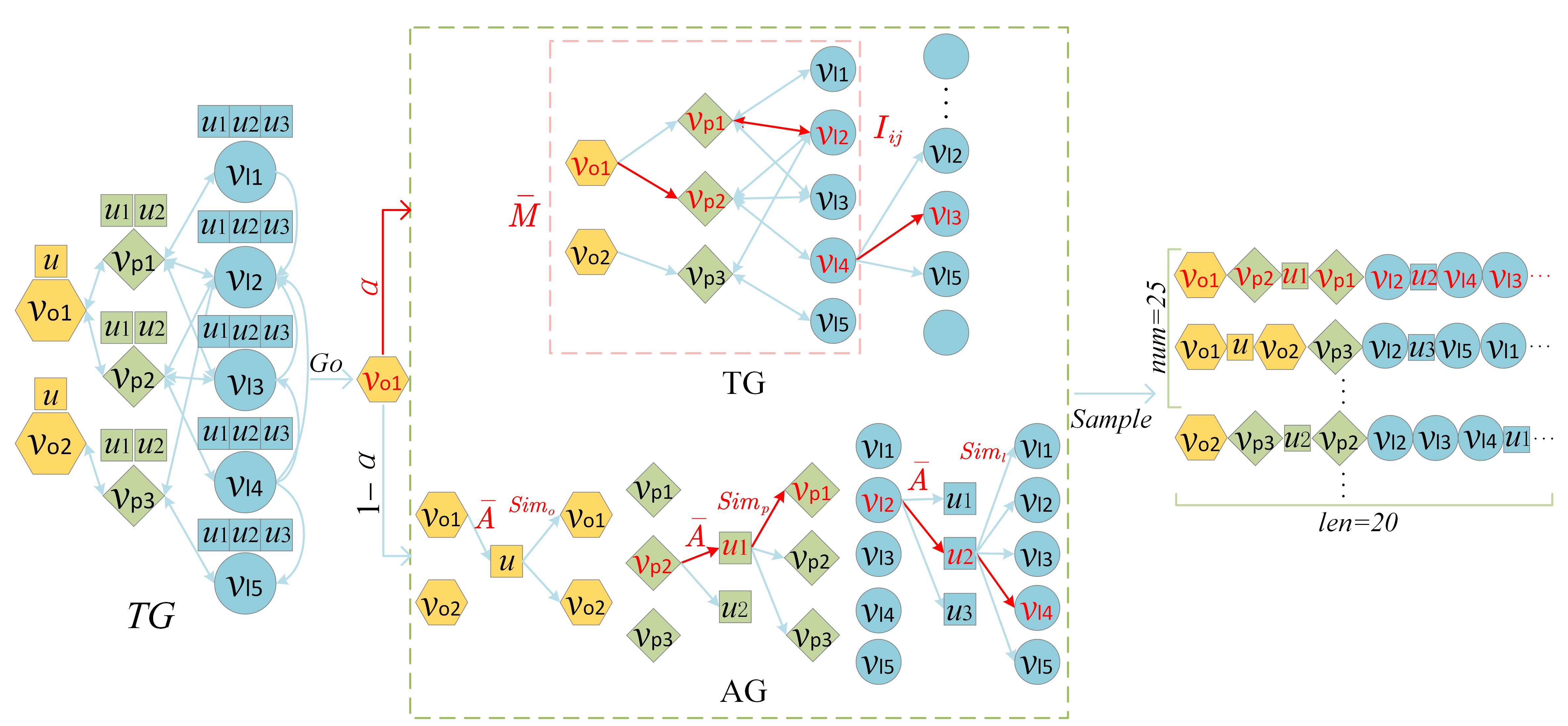}
    \caption{\revv{Joint random walk sampling procedure. Joint sampling is performed over the Trip Graph (TG) and multiple Attribute-Guided Graphs (AGs). The mixing coefficient $\alpha$ controls the probability of switching between TG-based and attribute-guided transitions at each step. Red arrows show an example sampling trajectory. On the right, $num$ denotes the number of sequences sampled per node, and $len$ indicates the fixed sequence length.}}
    \label{fig:g2}
\end{figure}

This subsection introduces HetGWalk, a joint random walk algorithm designed to generate heterogeneous node sequences by sampling over both the Trip Graph (TG) and the Attribute-guided Graphs (AGs). Walks on the TG exploit OD–path–segment relationships to capture OD-induced functional dependencies together with the underlying network topology, while walks on the AGs leverage node–attribute associations to encode functional similarities among nodes of the same type. By enriching node context with attribute information, the AGs increase sequence diversity and alleviate the tendency of random walks to oversample high-centrality nodes. By integrating these two complementary sampling mechanisms, HetGWalk produces context-rich heterogeneous sequences that jointly encode functional and physical structures, providing informative inputs for downstream representation learning. Figure \ref{fig:g2} illustrates the overall sampling procedure.

Each walk starts from an OD node and probabilistically traverses heterogeneous nodes in the TG as well as the three AGs corresponding to OD pairs, paths, and road segments. To jointly exploit the local and global functional structures encoded in the TG, HetGWalk integrates depth-first and breadth-first exploration strategies to balance OD-consistent path continuity and cross-path, cross-OD exploration. At each step, given the current node $v^{\phi}_i$, where $\phi \in \left\{o,p,l \right\}$ denotes the node type (OD pair, path, or road segment), a Bernoulli trial with probability $\alpha$ determines whether the next transition is sampled from the TG or from the AGs. When sampling on the TG, the transition probability is defined as

\begin{equation}
    P_{i, j}^{T G}=\beta \tilde{P}_{i, j}^{T G}+(1-\beta) \tilde{P}_{i^{-}, j}^{T G} \label{eq:e2}
\end{equation}
where $\beta$ controls the trade-off between depth-first and breadth-first exploration, $i^{-}$ denotes the predecessor of node $v^{\phi}_i$ in the walk, and $\tilde{P}_{i, j}^{TG}$ represents the depth-first transition probability on the TG. This probability depends on the types of the current and next nodes and is calculated as

\begin{equation}
\tilde{P}_{i, j}^{T G}=\left[\begin{array}{ccc}
0 & M^{o,p} & 0 \\
0 & 0 & \bar{M}^{p,l} \\
0 & (1-\varepsilon) \bar{M}^{\prime p,l} & \varepsilon \bar{M}^{l,l}
\end{array}\right] \in \mathbb{R}^{|V| \times|V|}
    \label{eq:e3}
\end{equation}
\revv{where $\bar{M}^{p,l}$ and $\bar{M}^{\prime p,l}$ are the row- and column-normalized path–segment incidence matrices, respectively, and $\bar{M}^{l,l}$ is the row-normalized segment–segment connectivity matrix defined in Definition 4.} The parameter $\varepsilon \in [0,1]$ biases the walk toward segment–segment transitions along OD-induced downstream dependencies, while $1-\varepsilon$ encourages returning to path nodes to preserve OD–path–segment consistency.

When the walk transitions to an AG, it follows an entity–attribute–entity pattern. First, the walk moves from the current entity node $v^{\phi}_i$ to one of its associated attribute nodes $u^{\phi}_k$. The transition probability $P_{i,k}^{AG_{\phi}}$ is proportional to the importance of attribute $u^{\phi}_k$ to node $v^{\phi}_i$, and is defined as the normalized attribute value $\bar{a}^{\phi}_{k,i}$:

\begin{equation}
P_{i,k}^{AG_{\phi}}=\bar{a}^{\phi}_{k,i}
    \label{eq:e4}
\end{equation}

This design ensures that attributes contributing more significantly to node $v^{\phi}_i$ are sampled with higher probability. Next, the walk transitions from the selected attribute node $u^{\phi}_k$ to a new entity node $v^{\phi}_j$. The corresponding transition probability $P_{k,j}^{AG_{\phi}}$ is determined by the functional similarity between nodes $v^{\phi}_i$ and $v^{\phi}_j$ with respect to attribute $u^{\phi}_k$, defined as:

\begin{equation}
    P_{k,j}^{AG_{\phi}}=sim_{i,j}^k=1-\left| \bar{a}^{\phi}_{k,j}-\bar{a}^{\phi}_{k,i} \right|
    \label{eq:e5}
\end{equation}

\begin{algorithm}[htb]
\SetAlgoNlRelativeSize{-2}
\caption{\textbf{HetGWalk}}
\label{alg:HetGWalk}
\KwIn{TG, $AG_o, AG_p, AG_l$, walk length $len$, walks per OD $num$}
\KwOut{The sampling sequence collections of $|V^o|$ nodes}

$S \leftarrow \emptyset$ \;
\For{each OD node $v^o_i \in V^o$} {
    \For{$n = 1$ \KwTo $num$} {
        $s \leftarrow [v^o_i]$, predecessor $v^{-} \leftarrow \text{null}$ \;
        \For{$t = 2$ \KwTo $len$} {
            current node $v^\phi_i \leftarrow s[t-1]$ with type $\phi$ \;
            Sample $z \sim \text{Bernoulli}(\alpha)$ \;
            \If{$z = 1$}{
                Sample $v^\phi_{j}$ from TG using $P_{i,j}^{TG}$ (Eq. (\ref{eq:e3})) given ($v^\phi_{i}, v^{-}$) \;
            }
            \Else{
                Sample attribute $u^{\phi}_k$ using $P^{AG_{\phi}}_{i,k}$ (Eq. (\ref{eq:e4})) \;
                Append $u^{\phi}_k$ to $s$ \;
                Sample $v^\phi_{j}$ using $P^{AG_{\phi}}_{k,j}$ (Eq. (\ref{eq:e5})) \;
                Append $v^\phi_{j}$ to $s$ \;
            }
            $v^{-} \leftarrow v^\phi_{i}$ \;
        }
        Add $s$ to $S$ \;
    }
}
\Return{$S$} \;
\end{algorithm}

The resulting similarity scores are normalized over candidate entity nodes to form a valid transition distribution. This formulation encourages the walk to transition to entity nodes that are functionally similar to the origin node $v^{\phi}_i$ under the selected attribute, even in the absence of direct connectivity in the original TG.

Overall, the probability of reaching node $v^{\phi}_j$ from $v^{\phi}_i$ via the AG is computed as a weighted aggregation over all possible attribute-mediated paths. Through this mechanism, the walk implicitly induces an attribute-aware similarity structure enriched with functional semantic, enabling the model to capture functional relationships beyond explicit graph connectivity. The implementation details of HetGWalk are described in Algorithm \ref{alg:HetGWalk}.

Lines 1–2 initialize an empty sequence set and iterate over all OD nodes. Lines 3–4 generate $num$ walk sequences for each OD node and initialize each walk with the OD node as the starting point. Lines 5–6 iteratively extend the walk by identifying the current node at each step. Line 7 samples a Bernoulli random variable to decide whether the next transition is taken on the TG or on an AG. Lines 8–9 sample the next node from the TG according to Eq. (\ref{eq:e2}). Lines 10–13 sample the next node via the corresponding AG using an entity–attribute–entity transition. Lines 14–15 append the sampled node(s) to the current walk and update the predecessor. Lines 16 add the completed walk to the sequence set, and Line 17 outputs the set of generated heterogeneous walk sequences.

\subsubsection{\rev{Theoretical Properties of HetGWalk}}
In this subsection, we formalize the properties of AGs and joint random walks on the TG and AGs for the proposed heterogeneous graph representation learning framework.

\rev{\textbf{Definition 7 }(\textit{Type-Specific Attribute Propagation Operators}): For a node $v_i^{\phi} \in V^{\phi}$, $\phi \in \left\{ o, p, l \right\}$, the propagation operator $\vartheta_i^{\phi}(j)$ measures the probability of influence propagation from $v_i^{\phi}$ to another node $v_j^{\phi}$ via attributes:}

\begin{equation}
    \rev{\vartheta_i^\phi(j)=\sum_{k \in U^o} \bar{a}_{k, i}^\phi \cdot\left(1-\left|\bar{a}_{k, j}^\phi-\bar{a}_{k, i}^\phi\right|\right), \quad v_j^\phi \in V^\phi 
}
\end{equation}

\rev{This operator captures type-specific semantic dependencies within the attribute-guided graph.}

\rev{\textbf{Theorem 1} (Type-Aware Dynamic Semantic Propagation): For any node type $\phi \in \left\{ o, p, l \right\}$, attribute-guided random walk transitions on $AG_{\phi}$ implement a type-specific semantic propagation process. Specifically, when the walk is currently at node $v_i^{\phi}$, the total transition probability from $v_i^{\phi}$ to any node $v_j^{\phi}$ is exactly equal to the propagation operator $\vartheta_i^\phi$.}

\rev{\textbf{Proof}: By the AG walk rules, the probability of moving from $v_i^{\phi}$ to $v_j^{\phi}$ via an attribute node $u_k^{\phi}$ is:}

\begin{equation}
    \rev{P_{i, k, j}^{A G_\phi}=P_{i, k}^{A G_\phi} \cdot P_{k, j}^{A G_\phi}=\bar{a}_{k, i}^\phi \cdot\left(1-\left|\bar{a}_{k, j}^\phi-\bar{a}_{k, i}^\phi\right|\right)}
\end{equation}

\rev{Summing over all attribute nodes yields the total transition probability:}

\begin{equation}
    \rev{P_{i, j}^{A G_\phi}=\sum_{k \in U^\phi} P_{i, k, j}^{A G_\phi}=\sum_{k \in U^\phi} \bar{a}_{k, i}^\phi \cdot\left(1-\left|\bar{a}_{k, j}^\phi-\bar{a}_{k, i}^\phi\right|\right)=\vartheta_i^\phi(j)}
\end{equation}

\rev{\textbf{Implication}: This formalizes that AG-based random walks inherently encode type-specific semantic similarity among nodes, providing a principled mechanism to bias walks toward functionally relevant nodes.}

\rev{\textbf{Theorem 2} (Multi-graph Ergodicity).}

\rev{Consider the multi-graph system $G_{sys} \left\{ TG, AG_o, AG_p, AG_l \right\}$. Under the following assumptions: }

\rev{(1) TG Connectivity: TG is strongly connected across all nodes via OD–path–segment relationships.}

\rev{(2) AG internal connectivity: Each $AG_{\phi}$ is semantically connected, i.e., there exists a path between any two nodes in attribute space.}

\rev{(3) Cross-graph reachability: TG ensures that any node type can reach any other node type.}

\rev{(4) Mixture balance: The mixing coefficient $\alpha \in (0, 1)$ ensuring probabilistic switching between TG and AG walks.}

\rev{The joint random walk defines an ergodic Markov chain with a unique stationary distribution $\pi^*$.}

\rev{\textbf{Proof}: The joint transition matrix for the multi-graph random walk is:}

\begin{equation}
    \rev{P^{joint}=\alpha P^{TG}+(1-\alpha) \sum_{\phi \in\{o, p, l\}} \delta_\phi P^{A G_\phi}}
\end{equation}

\rev{where $\delta_\phi=1$ if the current node is of type $\phi$ and 0 otherwise.}

\rev{\textbf{Irreducibility}: TG connectivity ensures any node type can reach any other type via TG. AG internal connectivity guarantees that nodes of the same type are mutually reachable through attributes. Thus, any pair of nodes in $V_{sys}$ can be connected via a finite sequence of transitions, ensuring irreducibility.}

\rev{\textbf{Aperiodicity}: Since $\alpha \in (0,1)$, the walk probabilistically switches between TG and AG, preventing fixed cycles. There exists a non-zero probability of returning to the current node (self-loop via AG), guaranteeing aperiodicity.}

\rev{Positive The state space $V_{sys}$ is finite, and the chain is irreducible. Hence, every state is positive recurrent.}

\rev{By the standard Markov chain theory, an irreducible, aperiodic, and positive recurrent finite chain is ergodic and possesses a unique stationary distribution $\pi^*$.}

\rev{\textbf{Implication}: Regardless of the starting node, sufficiently long joint walks produce sequences whose sampling distribution converges to $\pi^*$, guaranteeing stability and repeatability of the embeddings learned from these sequences.}

\textbf{Discussion}: The two theorems collectively demonstrate that TG and AGs provide complementary guidance for heterogeneous random walks. Specifically, the multiple AGs enable type-specific modeling of functional similarities, allowing transition probabilities to capture distinct functional semantics—such as OD-flow intensity, path-level relationships, and segment attributes. Meanwhile, the TG preserves multi-hop, cross-type functional dependencies, ensuring that walks capture essential functional-structural relationships among OD pairs, paths, and segments. Together, the TG and AGs form a coordinated mechanism that biases walks toward nodes that are both semantically relevant and structurally dependent. \revv{This coordination directly affects the distribution and context of sampled nodes, increasing the exposure of functionally critical segments during training. As a result, segments that are important under specific OD flows or path interactions are embedded in richer and more informative contexts, yielding representations that are better suited for downstream ranking.}

\subsection{Road segment embedding}
\rev{In the heterogeneous sequences generated by HetGWalk, road segments, OD pairs, paths, and their associated attributes are interleaved in a way that can induces long-range contextual dependencies. Traditional recurrent architectures such as GRUs \citep{gru} or LSTMs \citep{lstm} rely on sequential inductive biases and may be less effective in modeling interactions between elements that are far apart in the sequence, especially when heterogeneous entities participate in multi-hop, cross-type relationships. To capture such long-range contextual dependencies, we adopt a Transformer encoder \citep{transformer}, whose self-attention mechanism enables direct modeling of pairwise interactions within each sequence without relying on positional proximity.}

\rev{Before feeding a sequence into the encoder, each node is linearly projected into a type-specific embedding space, ensuring that heterogeneous entities (OD pairs, paths, segments, and attributes) are represented within a unified latent domaing. Multiple self-attention layers progressively integrate contextual information, producing context-dependent embeddings for each element. For a road segment $v_i^l$ appearing in multiple sequences, the encoder generates multiple embeddings reflecting its varying context. These embeddings are aggregated using an attention-based multiple instance learning (AMIL) mechanism \citep{AMIL}, yielding a final segment representation $h^i$ that adaptively weights each contextual embedding according to its relative importance. }

\begin{equation}
\rev{ h^{i} = \sum_{b=1}^{\mathcal{B}} \frac{\exp\!\left\{ W^{\mathrm{T}}_{\mathcal{B}1}\tanh\!\big( W_{\mathcal{B}2} h^{l}_{i,b} \big) \right\}}
{\sum_{j=1}^{\mathcal{B}} \exp\!\left\{ W^{\mathrm{T}}_{\mathcal{B}1}\tanh\!\big( W_{\mathcal{B}2} h^{l}_{i,j} \big) \right\}}\, h^{l}_{i,b} } 
\end{equation}
\rev{where $\mathcal{B}$ denotes the number of embeddings obtained by the transformer encoder for $v_i^l$, $W_{\mathcal{B}1}$ and $W_{\mathcal{B}2}$ are learnable parameters.}

\rev{The multi-head self-attention mechanism allows the model to capture co-occurrence patterns and functional relationships among road segments from multiple perspectives. When segments frequently co-occur within the same sequences, different attention heads can capture diverse types of dependencies, including spatial interactions associated with shared OD flows or related paths, as well as attribute-based similarities such as traffic volume, lane count, and speed category. This parallel processing facilitates the formation of semantically coherent segment representations in the embedding space. As a result, the final embedding for each segment integrates topological context, OD-induced long-range dependencies, and attribute-based similarities, yielding a context-aware representation suitable for assessing segment criticality.}

\subsection{Listwise learning to rank module}
We designed a listwise ranking module that predicts node importance rankings \rev{from the learned embeddings.} This module consists of two stacked fully-connected layers, with a Dropout layer after the first layer to enhance generalization. \rev{It takes a list of $K$ road segment embeddings as input and outputs a list of predicted ranking scores.}

\rev{To optimize this module, we adopt the well-established ListNet framework \citep{listnet}, specifically its efficient Top-One probability variant, as our training objective. This choice is motivated by its superior alignment with our goal of identifying the most critical segments. The loss function is defined as the Kullback-Leibler (KL) divergence between the ground-truth and predicted probability distributions, which are obtained by applying the SoftMax function to the ground-truth importance scores and the model's predicted scores, respectively:}

\begin{equation}
L(Z) = D_{KL}(Y_Z \parallel \hat{Y}_Z) = \sum_{v_j \in Z} y_j \log \left( \frac{y_j}{\hat{y}_j} \right)
    \label{eq:e11}
\end{equation}
where $\hat{y}_j$ and $y_j$ denote the predicted ranking score and ground-truth ranking label, respectively, for the \textit{j}-th road segment node in the list $Z$.

\rev{Intuitively, this loss function emphasizes the accuracy at the top of the ranking. It heavily penalizes the model if it assigns a low probability to a truly critical segment, thereby ensuring that the learning process is dominated by the most important examples.}

\subsection{\rev{Computational Complexity Analysis}}
\rev{This subsection analyzes the time complexity of HetGL2R and compares it with several  representative baseline methods.}

\subsubsection{\rev{Complexity Analysis of HetGL2R}}
\revv{The construction of the heterogeneous trip graph is performed once in an offline manner and scales linearly with the number of OD pairs, paths, and road segments, without requiring iterative optimization. Consequently, the overall computational cost of HetGL2R is dominated by the following three core components:}

\begin{itemize}
\setlength{\itemsep}{0pt}
    \item \rev{HetGWalk Sampling Module: The overall computational complexity of the joint random-walk procedure consists of a one-time preprocessing stage and the subsequent walk-generation stage. In preprocessing, Alias tables are constructed for all nodes in the Trip Graph (TG) and the Attribute-Guided Graphs (AGs), incurring a total cost of $O(m+\sum_{\phi \in\{o, p, l\}} a^\phi)$, where $m$ is the number of TG edges and $a^\phi$ is the number of attribute edges in $AG_{\phi}$; The memory requirement is of the same asymptotic order. After preprocessing, each TG step requires one Alias lookup, and each AG step (entity–attribute–entity) requires two, all in $O(1)$ time. Thus, generating $num$ walks of length $len$ from each of the $n$ nodes costs $O(num \cdot n \cdot len)$ in total. Overall, the joint random-walk procedure requires $O(m+\sum_{\phi \in\{o, p, l\}} a^\phi+num \cdot n \cdot len)$, where the first two terms correspond to preprocessing and the last term accounts for walk generation.} \revv{All heterogeneous random walks are generated with fixed length and bounded sampling budgets, and rely only on local transitions. As a result, the walk generation process scales linearly with the number of sampled walks and can be efficiently parallelized.}

    \item \rev{Transformer Encoder Module: The sequences generated by HetGWalk are fed into a Transformer encoder to learn context-aware node embeddings. For sequences of length $len$ with hidden dimension $d$, the self-attention computation has complexity $O(len^2 \cdot d)$ per sequence, and the feed-forward network contributes $O(len^2 \cdot d)$. If $num$ walks are generated for each of the $n$ nodes, the total per-epoch cost is $O(num \cdot n \cdot len^2 \cdot d+num \cdot n \cdot len \cdot d^2)$, with memory requirements of $O(len^2)$ for attention maps and $O(len \cdot d)$ for embeddings per sequence.} \revv{Notably, the Transformer encoder is applied to sampled walk sequences rather than the entire heterogeneous graph. Therefore, its computational cost depends on the sequence length instead of the global network size, avoiding the quadratic complexity associated with full-graph attention.}

    \item \rev{Listwise Ranking Module: This module consists of a simple feed-forward neural network with a complexity of $O(K \cdot d^2)$, where $K$ is the ranking list length. The computational cost of this module is negligible compared to the preceding steps.}
\end{itemize}

\revv{From an implementation perspective, the main components of HetGL2R exhibit inherent parallelism. Random walk sampling and sequence encoding are performed on independent samples, while the listwise ranking stage introduces minimal additional overhead. As a result, the overall complexity of HetGL2R is linear with respect to the number of sampled sequences and quadratic with respect to the sequence length, supporting efficient training and inference in large-scale settings using standard parallel computing resources.}

\subsubsection{\rev{Complexity Comparison with Baseline Methods}}
\rev{We compare HetGL2R with two main categories of baseline methods:}

\begin{itemize}
\setlength{\itemsep}{0pt}
    \item \rev{Message-Passing GNNs (e.g., HGT \citep{rf48}, GAT \citep{rf50}). These models learn node representations through iterative neighborhood aggregation. Each layer incurs a computational cost of $O(|E|d)$, where $|E|$ denotes the number of edges and d the feature dimensionality. Modeling long-range dependencies requires stacking multiple layers, causing the total complexity to increase with both the graph size and the network depth. In addition, the layer-wise dependency inherent to message passing introduces sequential execution constraints, which fundamentally limit the degree of parallelism achievable within a single batch.}

    \item \rev{Random Walk-based Homogeneous Methods (e.g., DeepWalk \citep{rf39}): The sampling stage of these methods shares a similar complexity with our HetGWalk, i.e., $O(num \cdot n \cdot len)$. The key computational difference emerges in the embedding learning phase. These methods typically employ a separate Skip-gram model, which, while highly efficient per training sample due to its shallow architecture and negative sampling, operates on a corpus of generated sequences in a distinct, non-end-to-end step. In contrast, our Transformer encoder, despite having a higher per-sequence complexity of $O(len^2 \cdot d)$, is integrated into an end-to-end framework. This design eliminates the need for a  separate training pipeline and allows for the direct co-optimization of all components, although this integration introduces a higher computational cost per model update.}
\end{itemize}

\rev{The computational profile of HetGL2R differs significantly from these baselines. Compared to multi-layer GNNs, our approach trades the iterative multi-hop aggregation for a more direct, sequence-based encoding of long-range contexts. Compared to  pipeline-based random walk methods, it consolidates the entire learning process into a single, unified optimization, which simplifies the training workflow at the cost of increased per-pass complexity. The highly parallelizable nature of both the random walk sampling and the sequence encoding stages helps mitigate this cost in practice.}

\section{Experiments}
HetGL2R is implemented with PyTorch. All of the experiments are conducted under a computer environment equipped with an AMD Ryzen 7 6800H processor, 16 GB of RAM, and an NVIDIA GeForce RTX 3050 Laptop GPU.

\subsection{\rev{Dataset Construction}}
\begin{table}[t]
  \centering
  \tablesize
  \caption{\rev{Description of the three synthetic datasets.}}
  \label{tab:t1}
  \setlength{\tabcolsep}{18pt} 
  \begin{tabular}{ccccc}
  \toprule \hline
dataset & od  & path  & road & cars  \\ \midrule
\rev{SY-Net110}      & 20  & 1762  & 110  & 2313  \\ 
\rev{SY-Net514}      & 90  & 5148  & 514  & 6334  \\ 
\rev{RD-Net3478}      & \rev{800}  & \rev{52579}  & \rev{3478}  & \rev{74174}  \\ 
 \hline  \bottomrule
  \end{tabular}
\end{table}

\rev{To evaluate the proposed framework, three traffic network datasets were constructed using the SUMO \citep{rf2} microscopic traffic simulator. Two datasets, SY-Net110 and SY-Net514, were generated from subregions of the real Shenyang urban road network, containing 110 and 514 nodes, respectively. These networks were extracted from OpenStreetMap and calibrated with realistic road capacities and speed limits. Moreover, SY-Net110 and SY-Net514 incorporated real peak-hour traffic flow data, from which the corresponding OD flows were inferred (20 OD pairs for SY-Net110 and 90 OD pairs for SY-Net514). In addition, a large-scale randomly generated road network, RD-Net3478, with 3,478 nodes and 800 OD pairs, was constructed to evaluate the scalability of the proposed model on a network with a distinct structural topology. Table \ref{tab:t1} summarizes the statistical properties of the three networks.}

To obtain a ground-truth ranking of road segment importance, we conducted simulation-based failure experiments using SUMO by reducing the maximum speed limit of each segment to 10\% of its original value. The number of other segments that subsequently experienced failure—operationally defined as a speed drop below 10\% of the corresponding speed limit—was recorded over multiple time windows. The final importance score (IS \citep{rf8}) for each segment was computed as a weighted sum of the affected segments across time, capturing the cascading impact of a segment failure on overall network performance.

\begin{equation}
IS_i = \sum_{t=1}^{T} \gamma^t n_t
    \label{eq:e12}
\end{equation}
where $\gamma$ is the decay coefficient (set to 0.9 in this paper), $n_t$ denotes the number of failed road segments in the network at time step $t$, and $T$ represents the total number of time steps.

\subsection{Baselines and evaluation metrics}
To evaluate the effectiveness of the proposed method, we compare it with a broad set of baselines, which we group into two main categories: learning-based methods and rule-based methods.

The learning-based baselines are further divided into two types:

\textbf{(1)} Representation learning (embedding) models that focus on learning node representations from graph topology, attributes, or language model prompts. These include DeepWalk \citep{rf39}, GraphRNA \citep{rf40}, TraNode2vec \citep{rf7}, HGT \citep{rf48}, Dir-GNN \citep{rf53}, GAT \citep{rf50}, GATv2 \citep{gatv2}, PDFormer \citep{PDFormer}, DR-HGNN \citep{DR-HGN}, SeHGNN \citep{SeHGNN}, Edge-GNN \citep{Edge-GNN}, ENGINE \citep{rf51}, and \textnormal{GNN\textsubscript{LLMEmb}} \citep{rf52}. For ENGINE and \textnormal{GNN\textsubscript{LLMEmb}}, structured attributes are converted into textual descriptions to meet language model input requirements.

\textbf{(2)} Specialized learning-to-rank (LTR) models that are explicitly designed for ranking tasks. These include MGL2Rank \citep{rf8}, LambdaLoss \citep{LambdaLoss}, RankFormer \citep{Rankformer}, and LightGBM \citep{lighrgbm}.

The rule-based baselines rank nodes based on topology- and attribute-driven heuristics without learnable parameters. These include Improved PageRank (IPageRank) \citep{rf3}, CRRank \citep{rf1}, and OVED-Rank \citep{rf44}.

To ensure a fair comparison, the node embeddings produced by all models in group \textbf{(1)} are fed into our proposed listwise LTR module to generate the final rankings.

\begin{table}[t]
  \caption{\rev{Ranking Performance of Different Methods on SY-Net110.}}
  \label{tab:sy110}
  \centering
  \tablesize
  \begin{tabular}{c|c|c c c c}
    \toprule\hline
    \multicolumn{2}{@{\hspace{2mm}}l|}{\textbf{\rev{SY-Net110}}} & \rev{EMD} & \rev{NDCG@K} & \rev{Diff} & \rev{$\tau$} \\
    \cline{1-6}
    \multirow{17}{*}{\rev{Learning-based}} & \rev{DeepWalk}         & \rev{0.3146} & \rev{0.7099} & \rev{0.4329} & \rev{0.5118} \\
                                & \rev{GraphRNA}         & \rev{0.1350} & \rev{0.8145} & \rev{0.3037} & \rev{0.6083} \\
                                & \rev{TraNode2vec}      & \rev{0.2154} & \rev{0.8011} & \rev{0.3203} & \rev{0.6004} \\
                                & \rev{MGL2Rank}         & \rev{0.1520} & \rev{0.8970}  & \rev{0.2969} & \rev{0.7311} \\
                                & \rev{HGT}              & \rev{0.1542} & \rev{0.9287} & \rev{0.2998} & \rev{0.5688} \\
                                & \rev{Dir-GNN}          & \rev{0.1635} & \rev{0.8847} & \rev{0.3152} & \rev{0.5330}  \\
                                & \rev{GAT}              & \rev{0.2540} & \rev{0.8757} & \rev{0.3004} & \rev{0.6444} \\
                                & \rev{ENGINE}           & \rev{0.2273} & \rev{0.8395} & \rev{0.3116} & \rev{0.5984} \\
                                & \rev{\textnormal{GNN\textsubscript{LLMEmb}}}
                                                        & \rev{0.1975} & \rev{0.8617} & \rev{0.3240}  & \rev{0.6246} \\
                                & \rev{GATv2}            & \rev{0.0889} & \rev{0.9073} & \rev{0.2874} & \rev{0.7422} \\
                                & \rev{PDFormer}         & \rev{0.0876} & \rev{0.8999} & \rev{0.2772} & \rev{0.7023} \\
                                & \rev{DR-HGNN}          & \rev{0.0774} & \rev{0.9013} & \rev{0.2685} & \rev{0.6146} \\
                                & \rev{SeHGNN}           & \rev{0.0977} & \rev{0.9057} & \rev{0.2846} & \rev{0.7592} \\
                                & \rev{Edge-GNN}         & \rev{0.1216} & \rev{0.9133} & \rev{0.2799} & \rev{0.7805} \\
                                & \rev{LambdaLoss}       & \rev{0.0974} & \rev{0.8463} & \rev{0.3469} & \rev{0.6413} \\
                                & \rev{RankFormer}       & \rev{0.2062} & \rev{0.7656} & \rev{0.4156} & \rev{0.5325} \\
                                & \rev{LightGBM}         & \rev{0.1502} & \rev{0.8347} & \rev{0.3597} & \rev{0.6557} \\
    \cline{1-6}
    \multirow{3}{*}{\rev{Rule-based}} & \rev{IPageRank}        & \rev{0.3083} & \rev{0.6266} & \rev{0.4658} & \rev{0.4705} \\
                                & \rev{CRRank}           & \rev{0.2144} & \rev{0.7797} & \rev{0.3659} & \rev{0.6239} \\
                                & \rev{OVED-Rank}        & \rev{0.3360} & \rev{0.7578} & \rev{0.3937} & \rev{0.6149} \\
    \cline{1-6}
    \rev{}                      & \rev{HetGL2R}          & \textbf{\rev{0.0699}} & \textbf{\rev{0.9459}} & \textbf{\rev{0.2483}} & \textbf{\rev{0.7934}} \\
    \hline\bottomrule
  \end{tabular}
  \caption*{\footnotesize \revv{Note: Best results are highlighted in bold. Higher values indicate better performance for NDCG@K and Kendall’s $\tau$, while lower values are better for EMD and Diff.}}
\end{table}

We employ the following metrics to evaluate all methods:

\textbf{(1)} Normalized Discounted Cumulative Gain (NDCG@K) evaluates ranking quality by assigning higher weights to top-ranked elements through logarithmic discounting, thereby emphasizing the correctness of the most important items:

\begin{equation}
\rev{\operatorname{NDCG@K} = \frac{\sum_{j=1}^{n} \frac{2^{n-\hat{y}_i} - 1}{\log_2(j+1)}}{\sum_{j=1}^{n} \frac{2^{n-y_i} - 1}{\log_2(j+1)}}}
    \label{eq:e13}
\end{equation}
\rev{where, $n$ denotes the number of nodes, and $y_i$ represents the sorted list.}

\textbf{(2)} Wasserstein distance (Earth Mover’s Distance, EMD) quantifies the discrepancy between the predicted and ground-truth importance distributions, capturing differences in both ranking order and magnitude.

\textbf{(3)} Ideal ranking list deviation (Diff) \citep{rf1} measures the relative deviation between the predicted ranking and the ideal ranking at each list position, reflecting position-sensitive ranking errors:

\begin{equation}
\rev{\operatorname{Diff}(Y,\hat{Y})=\frac{\sum_{i=1}^n\left|\mathcal{P}\mathit{os}\left(v_i^l, Y\right)-\mathcal{P}\mathit{os}\left(v_i^l, \hat{Y}\right)\right|}{\left\lfloor n^2 / 2\right\rfloor}}
    \label{eq:e14}
\end{equation}
\rev{where $\mathcal{P}\mathit{os}\left(v_i^l, Y\right)$ returns the order number of $v_i^l$ in the list $Y$.}

\textbf{(4)} Kendall’s rank correlation coefficient (Kendall’s $\tau$) \citep{tau} assesses the ordinal consistency between two rankings by comparing the numbers of concordant and discordant element pairs:

\begin{equation}
    \rev{\tau\!\left(Y,\hat{Y}\right)
= \frac{\operatorname{count}_{c}-\operatorname{count}_{d}}{0.5\,n\,(n-1)}}
\end{equation}
\rev{where $\operatorname{count}_{c}$ and $\operatorname{count}_{d}$ denote the number of consistent and inconsistent pairs, respectively. The coefficient ranges from $\tau=1$ (perfect agreement) to $\tau=-1$ (perfect disagreement), with $\tau=0$ indicating no correlation. Unlike $\operatorname{NDCG@K}$, which focuses on the top-K items, Kendall’s $\tau$ evaluates the global consistency of the entire ranking.}

\begin{table}[t]
  \caption{\rev{Ranking Performance of Different Methods on SY-Net514.}}
  \label{tab:sy514}
  \centering
  \tablesize
  \begin{tabular}{c|c|c c c c}
    \toprule\hline
    \multicolumn{2}{@{\hspace{2mm}}l|}{\textbf{\rev{SY-Net514}}} & \rev{EMD} & \rev{NDCG@K} & \rev{Diff} & \rev{$\tau$} \\
    \cline{1-6}
    \multirow{17}{*}{\rev{Learning-based}} & \rev{DeepWalk}         & \rev{0.2106} & \rev{0.7413} & \rev{0.3643} & \rev{0.5005} \\
                                & \rev{GraphRNA}         & \textbf{\rev{0.1103}} & \rev{0.8572} & \rev{0.3241} & \rev{0.5924} \\
                                & \rev{TraNode2vec}      & \rev{0.1822} & \rev{0.8599} & \rev{0.3274} & \rev{0.5680} \\
                                & \rev{MGL2Rank}         & \rev{0.1357} & \rev{0.9067} & \rev{0.2985} & \rev{0.7473} \\
                                & \rev{HGT}              & \rev{0.1309} & \rev{0.8440} & \rev{0.3290} & \textbf{\rev{0.8961}} \\
                                & \rev{Dir-GNN}          & \rev{0.1381} & \rev{0.8243} & \rev{0.3886} & \rev{0.7037} \\
                                & \rev{GAT}              & \rev{0.1371} & \rev{0.8687} & \rev{0.3001} & \rev{0.6905} \\
                                & \rev{ENGINE}           & \rev{0.1723} & \rev{0.8623} & \rev{0.3109} & \rev{0.5934} \\
                                & \rev{\textnormal{GNN\textsubscript{LLMEmb}}}
                                                        & \rev{0.1840} & \rev{0.8759} & \rev{0.2988} & \rev{0.6012} \\
                                & \rev{GATv2}            & \rev{0.2195} & \rev{0.8754} & \rev{0.3070} & \rev{0.6901} \\
                                & \rev{PDFormer}         & \rev{0.1956} & \rev{0.9189} & \rev{0.2659} & \rev{0.8226} \\
                                & \rev{DR-HGNN}          & \rev{0.1964} & \rev{0.9118} & \rev{0.2693} & \rev{0.7893} \\
                                & \rev{SeHGNN}           & \rev{0.1961} & \rev{0.9007} & \rev{0.2946} & \rev{0.7491} \\
                                & \rev{Edge-GNN}         & \rev{0.1952} & \rev{0.9163} & \rev{0.2592} & \rev{0.8007} \\
                                & \rev{LambdaLoss}       & \rev{0.1799} & \rev{0.8840} & \rev{0.3201} & \rev{0.6915} \\
                                & \rev{RankFormer}       & \rev{0.1942} & \rev{0.8685} & \rev{0.3624} & \rev{0.6178} \\
                                & \rev{LightGBM}         & \rev{0.2240} & \rev{0.8890} & \rev{0.3094} & \rev{0.7390} \\
    \cline{1-6}
    \multirow{3}{*}{\rev{Rule-based}} & \rev{IPageRank}        & \rev{0.2443} & \rev{0.6573} & \rev{0.4446} & \rev{0.4857} \\
                                & \rev{CRRank}           & \rev{0.1781} & \rev{0.7464} & \rev{0.4089} & \rev{0.5275} \\
                                & \rev{OVED-Rank}        & \rev{0.2833} & \rev{0.7809} & \rev{0.3967} & \rev{0.5890} \\
    \cline{1-6}
    \rev{}                      & \rev{HetGL2R}          & \rev{0.1318} & \textbf{\rev{0.9471}} & \textbf{\rev{0.2478}} & \rev{0.8898} \\
    \hline\bottomrule
  \end{tabular}
\end{table}

\begin{table}[t]
  \caption{\rev{Ranking Performance of Different Methods on RD-Net3478.}}
  \label{tab:rd3478}
  \centering
  \tablesize
  \begin{tabular}{c|c|c c c c}
    \toprule\hline
    \multicolumn{2}{@{\hspace{2mm}}l|}{\textbf{\rev{RD-Net3478}}} & \rev{EMD} & \rev{NDCG@K} & \rev{Diff} & \rev{$\tau$} \\
    \cline{1-6}
    \multirow{17}{*}{\rev{Learning-based}} & \rev{DeepWalk}       & \rev{0.4638} & \rev{0.6525} & \rev{0.4850} & \rev{0.3114} \\
                                & \rev{GraphRNA}       & \rev{0.3209} & \rev{0.7545} & \rev{0.3843} & \rev{0.5229} \\
                                & \rev{TraNode2vec}    & \rev{0.5175} & \rev{0.7428} & \rev{0.3957} & \rev{0.4535} \\
                                & \rev{MGL2Rank}       & \rev{0.3938} & \rev{0.8066} & \rev{0.3492} & \rev{0.6629} \\
                                & \rev{HGT}            & \rev{0.4076} & \rev{0.7630} & \rev{0.3781} & \rev{0.5679} \\
                                & \rev{Dir-GNN}        & \rev{0.4114} & \rev{0.7405} & \rev{0.4042} & \rev{0.4567} \\
                                & \rev{GAT}            & \rev{0.4167} & \rev{0.7394} & \rev{0.4176} & \rev{0.3974} \\
                                & \rev{ENGINE}         & \rev{0.4180} & \rev{0.7610} & \rev{0.3923} & \rev{0.5348} \\
                                & \rev{\textnormal{GNN\textsubscript{LLMEmb}}}
                                                     & \rev{0.4197} & \rev{0.7751} & \rev{0.3987} & \rev{0.3854} \\
                                & \rev{GATv2}          & \rev{0.4043} & \rev{0.7904} & \rev{0.3742} & \rev{0.6524} \\
                                & \rev{PDFormer}       & \rev{0.3265} & \rev{0.8495} & \rev{0.3441} & \rev{0.6990} \\
                                & \rev{DR-HGNN}        & \rev{0.3504} & \rev{0.8196} & \rev{0.3679} & \rev{0.7361} \\
                                & \rev{SeHGNN}         & \rev{0.3517} & \rev{0.8214} & \rev{0.3796} & \rev{0.6817} \\
                                & \rev{Edge-GNN}       & \rev{0.3554} & \rev{0.8160} & \rev{0.3773} & \rev{0.7829} \\
                                & \rev{LambdaLoss}     & \rev{0.3476} & \rev{0.7858} & \rev{0.3842} & \rev{0.5836} \\
                                & \rev{RankFormer}     & \rev{0.3672} & \rev{0.7740} & \rev{0.4170} & \rev{0.5680} \\
                                & \rev{LightGBM}       & \rev{0.4686} & \rev{0.7552} & \rev{0.3769} & \rev{0.5323} \\
    \cline{1-6}
    \multirow{3}{*}{\rev{Rule-based}} & \rev{IPageRank}      & \rev{0.4431} & \rev{0.6029} & \rev{0.5035} & \rev{0.2254} \\
                                & \rev{CRRank}         & \rev{0.3802} & \rev{0.7591} & \rev{0.4074} & \rev{0.5006} \\
                                & \rev{OVED-Rank}      & \rev{0.3923} & \rev{0.6948} & \rev{0.4510} & \rev{0.4050} \\
    \cline{1-6}
    \rev{}                      & \rev{HetGL2R}        & \textbf{\rev{0.3208}} & \textbf{\rev{0.8502}} & \textbf{\rev{0.3318}} & \textbf{\rev{0.7918}} \\
    \hline\bottomrule
  \end{tabular}
\end{table}

For HetGWalk, we set the propagation factor $e$ to 2, the walk bias factor $\alpha$ to 0.6, the number of sampled sequences to 25, and the sequence length to 20. \revv{The encoder consists of 6 layers, each with 8 attention heads and an embedding dimension of 64. We optimize the model using the Adam optimizer with a learning rate of 0.001.} All experiments are conducted on the same synthetic dataset under these settings to ensure a fair comparison.

\begin{figure}[t]
    \centering
    \includegraphics[width=1\linewidth]{aNet.jpg}
    \caption{\revv{Performance comparison of different methods on SY-Net110. The horizontal axis denotes the predicted rank of road segments, while the vertical axis represents their ground-truth importance scores (IS). Each point corresponds to a road segment, and different colors indicate different baseline methods and the proposed HetGL2R.}}
    \label{fig:anet}
\end{figure}

\begin{figure}[htbp]
    \centering
    \includegraphics[width=1\linewidth]{bNet.jpg}
    \caption{\revv{The Performance of Each Method on SY-Net514.}}
    \label{fig:bnet}
\end{figure}

\begin{figure}[htbp]
    \centering
    \includegraphics[width=1\linewidth]{cNet.jpg}
    \caption{\revv{The Performance of Each Method on RD-Net3478.}}
    \label{fig:cnet}
\end{figure}

\subsection{\rev{Experimental results}}

\rev{For each network (SY-Net110, SY-Net514, and RD-Net3478), 70\% of the nodes were randomly selected for training, with the remaining 30\% reserved for testing. Tables \ref{tab:sy110} - \ref{tab:rd3478} report the ranking performance of all methods across multiple evaluation metrics. The results consistently show that learning-based ranking models outperform non-learning baselines across all datasets, while HetGL2R achieves the best overall performance. Specifically, HetGL2R improves $\operatorname{NDCG@K}$, Diff, and $\tau$ by at least 1.82\%, 7.52\%, and 1.63\% on SY-Net110, by 2.98\% and 4.40\% on $\operatorname{NDCG@K}$ and Diff on SY-Net514 (ranking second only to HGT in $\tau$, with a marginal 0.70\% gap), and by 3.39\%, 4.98\%, and 1.12\% on SY-Net3478. These improvements confirm that the proposed embedding module effectively captures both the structural and functional dependencies of road segments, leading to robust and accurate ranking results.}

\begin{table}[htbp]
  \caption{\rev{Predicted Rankings of the Top 10 Nodes in the Ground Truth of SY-Net110.}}
  \label{tab:top10-110}
  \centering
  \small
  \resizebox{\linewidth}{!}{%
\begin{tabular}{|ccccccccccc|}
\hline
\rev{DeepWalk} & \rev{GraphRNA} & \rev{TraNode2vec} & \rev{MGL2Rank} & \rev{HGT} & \rev{Dir-GNN} & \rev{GAT} & \rev{ENGINE} & \rev{\textnormal{GNN\textsubscript{LLMEmb}}} & \rev{GATv2} & \rev{PDFormer} \\ \hline
\rev{2}  & \rev{3}  & \rev{1}  & \rev{1}  & \rev{1}  & \rev{1}  & \rev{1}  & \rev{1}  & \rev{6}  & \rev{1}  & \rev{1}  \\
\rev{3}  & \rev{7}  & \rev{9}  & \rev{2}  & \rev{2}  & \rev{2}  & \rev{6}  & \rev{5}  & \rev{8}  & \rev{2}  & \rev{2}  \\
\rev{5}  & \rev{5}  & \rev{4}  & \rev{3}  & \rev{7}  & \rev{3}  & \rev{3}  & \rev{2}  & \rev{3}  & \rev{3}  & \rev{3}  \\
\rev{4}  & \rev{2}  & \rev{3}  & \rev{4}  & \rev{5}  & \rev{4}  & \rev{4}  & \rev{3}  & \rev{5}  & \rev{4}  & \rev{4}  \\
\rev{1}  & \rev{4}  & \rev{2}  & \rev{5}  & \rev{6}  & \rev{16} & \rev{2}  & \rev{7}  & \rev{7}  & \rev{5}  & \rev{5}  \\
\rev{8}  & \rev{8}  & \rev{8}  & \rev{6}  & \rev{3}  & \rev{8}  & \rev{10} & \rev{10} & \rev{4}  & \rev{6}  & \rev{6}  \\
\rev{11} & \rev{13} & \rev{19} & \rev{9}  & \rev{8}  & \rev{7}  & \rev{16} & \rev{12} & \rev{2}  & \rev{11} & \rev{9}  \\
\rev{7}  & \rev{6}  & \rev{6}  & \rev{7}  & \rev{10} & \rev{5}  & \rev{8}  & \rev{4}  & \rev{27} & \rev{7}  & \rev{7}  \\
\rev{19} & \rev{10} & \rev{31} & \rev{8}  & \rev{4}  & \rev{9}  & \rev{25} & \rev{32} & \rev{19} & \rev{8}  & \rev{8}  \\
\rev{6}  & \rev{1}  & \rev{10} & \rev{11} & \rev{20} & \rev{6}  & \rev{9}  & \rev{8}  & \rev{28} & \rev{10} & \rev{10} \\ \hline
\rev{DR-HGNN} & \rev{SeHGNN} & \rev{Edge-GNN} & \rev{LambdaLoss} & \rev{RankFormer} & \rev{LightGBM} & \rev{IPageRank} & \rev{CRRank} & \rev{OVED-Rank} & \rev{HetGL2R} & \rev{Ground-Truth} \\ \hline
\rev{1}  & \rev{1}  & \rev{1}  & \rev{1}  & \rev{1}  & \rev{2}  & \rev{2}  & \rev{1}  & \rev{1}  & \rev{1}  & \rev{1}  \\
\rev{3}  & \rev{2}  & \rev{2}  & \rev{2}  & \rev{2}  & \rev{1}  & \rev{4}  & \rev{2}  & \rev{2}  & \rev{2}  & \rev{2}  \\
\rev{2}  & \rev{3}  & \rev{3}  & \rev{6}  & \rev{3}  & \rev{8}  & \rev{1}  & \rev{3}  & \rev{3}  & \rev{3}  & \rev{3}  \\
\rev{4}  & \rev{4}  & \rev{4}  & \rev{3}  & \rev{4}  & \rev{5}  & \rev{5}  & \rev{4}  & \rev{4}  & \rev{4}  & \rev{4}  \\
\rev{5}  & \rev{10}  & \rev{5}  & \rev{4}  & \rev{5}  & \rev{7}  & \rev{3}  & \rev{5}  & \rev{7}  & \rev{5}  & \rev{5}  \\
\rev{6}  & \rev{13}  & \rev{6}  & \rev{12} & \rev{6}  & \rev{6}  & \rev{7}  & \rev{11}  & \rev{11}  & \rev{7}  & \rev{6}  \\
\rev{12} & \rev{7}  & \rev{9}  & \rev{19} & \rev{13} & \rev{4}  & \rev{21} & \rev{7}  & \rev{8}  & \rev{6}  & \rev{7}  \\
\rev{7}  & \rev{9}  & \rev{7}  & \rev{5}  & \rev{7}  & \rev{19} & \rev{9}  & \rev{10}  & \rev{10}  & \rev{8}  & \rev{8}  \\
\rev{9}  & \rev{8}  & \rev{8}  & \rev{9}  & \rev{11} & \rev{20} & \rev{16} & \rev{9} & \rev{9}  & \rev{10}  & \rev{9}  \\
\rev{11} & \rev{12} & \rev{10} & \rev{7}  & \rev{12} & \rev{12} & \rev{17} & \rev{12}  & \rev{14} & \rev{9} & \rev{10} \\ \hline
\end{tabular}
}
\end{table}

To provide an intuitive comparison of algorithm performance, Figures \ref{fig:anet} - \ref{fig:cnet} illustrate the correspondence between predicted rankings and ground-truth importance scores (IS) across the SY-Net110, SY-Net514, and RD-Net3478 datasets. Overall, as predicted rankings decrease, the ground-truth IS values for all methods exhibit the expected monotonic downward trend. In the two simulated real-world road networks, SY-Net110 and SY-Net514, HetGL2R exhibits the smoothest and most stable decline, with tightly clustered scatter points closely following the ground truth, indicating strong ranking consistency. Several learning-based baselines, including MGL2Rank, SeHGNN, PDFormer, and Edge-GNN, also demonstrate reasonable alignment, albeit with larger fluctuations. In the randomly generated RD-Net3478 network, performance differences among methods become more pronounced. Models that primarily rely on topological structure (e.g., DeepWalk, GAT, \textnormal{GNN\textsubscript{LLMEmb}}, and IPageRank) exhibit more scattered distributions and larger deviations from the true rankings, whereas HetGL2R maintains robust behavior with orderly and concentrated scatter distributions.

\begin{table}[!htbp]
  \caption{\rev{Predicted Rankings of the Top 10 Nodes in the Ground Truth of SY-Net514.}}
  \label{tab:top10-514}
  \centering
  \small
  \resizebox{\linewidth}{!}{%
\begin{tabular}{|ccccccccccc|}
\hline
\rev{DeepWalk} & \rev{GraphRNA} & \rev{TraNode2vec} & \rev{MGL2Rank} & \rev{HGT} & \rev{Dir-GNN} & \rev{GAT} & \rev{ENGINE} & \rev{\textnormal{GNN\textsubscript{LLMEmb}}} & \rev{GATv2} & \rev{PDFormer} \\ \hline
\rev{22}  & \rev{1}  & \rev{1}  & \rev{1}  & \rev{1}  & \rev{1}  & \rev{1}  & \rev{69}  & \rev{1}  & \rev{1}  & \rev{1}  \\
\rev{25}  & \rev{4}  & \rev{5}  & \rev{5}  & \rev{3}  & \rev{4}  & \rev{5}  & \rev{12} & \rev{2}  & \rev{3}  & \rev{5}  \\
\rev{1}   & \rev{3}  & \rev{2}  & \rev{2}  & \rev{2}  & \rev{3}  & \rev{2}  & \rev{1}  & \rev{7}  & \rev{2}  & \rev{2}  \\
\rev{8}   & \rev{2}  & \rev{4}  & \rev{4}  & \rev{4}  & \rev{2}  & \rev{4}  & \rev{6}  & \rev{69} & \rev{4}  & \rev{4}  \\
\rev{2}   & \rev{12} & \rev{3}  & \rev{3}  & \rev{5}  & \rev{5}  & \rev{3}  & \rev{2}  & \rev{33} & \rev{5}  & \rev{3}  \\
\rev{7}   & \rev{13} & \rev{6}  & \rev{6}  & \rev{6}  & \rev{6}  & \rev{6}  & \rev{5}  & \rev{4}  & \rev{6}  & \rev{6}  \\
\rev{55}  & \rev{7}  & \rev{9}  & \rev{12} & \rev{7}  & \rev{7}  & \rev{8}  & \rev{44} & \rev{10} & \rev{7}  & \rev{9}  \\
\rev{13}  & \rev{10} & \rev{7}  & \rev{7}  & \rev{8}  & \rev{9}  & \rev{7}  & \rev{8}  & \rev{9}  & \rev{8}  & \rev{7}  \\
\rev{45}  & \rev{9}  & \rev{12} & \rev{13} & \rev{9}  & \rev{11} & \rev{10}  & \rev{30} & \rev{14} & \rev{11} & \rev{10} \\
\rev{46}  & \rev{17} & \rev{14} & \rev{14} & \rev{10} & \rev{13} & \rev{11} & \rev{31} & \rev{3}  & \rev{10} & \rev{11} \\ \hline
\rev{DR-HGNN} & \rev{SeHGNN} & \rev{Edge-GNN} & \rev{LambdaLoss} & \rev{RankFormer} & \rev{LightGBM} & \rev{IPageRank} & \rev{CRRank} & \rev{OVED-Rank} & \rev{HetGL2R} & \rev{Ground-Truth} \\ \hline
\rev{1}   & \rev{1}  & \rev{2}  & \rev{2}  & \rev{14} & \rev{1}  & \rev{21} & \rev{3}  & \rev{1}  & \rev{1}  & \rev{1}  \\
\rev{3}   & \rev{4}  & \rev{5}  & \rev{6}  & \rev{24} & \rev{5}  & \rev{2}  & \rev{1}  & \rev{2}  & \rev{3}  & \rev{2}  \\
\rev{4}   & \rev{7}  & \rev{1}  & \rev{1}  & \rev{43} & \rev{3}  & \rev{51} & \rev{7}  & \rev{7}  & \rev{2}  & \rev{3}  \\
\rev{20}  & \rev{8}  & \rev{4}  & \rev{4}  & \rev{4}  & \rev{2}  & \rev{53} & \rev{4}  & \rev{9}  & \rev{4}  & \rev{4}  \\
\rev{2}   & \rev{6}  & \rev{3}  & \rev{3}  & \rev{51} & \rev{4}  & \rev{20} & \rev{2}  & \rev{4}  & \rev{5}  & \rev{5}  \\
\rev{5}   & \rev{2}  & \rev{6}  & \rev{5}  & \rev{17} & \rev{7}  & \rev{7}  & \rev{5}  & \rev{5}  & \rev{6}  & \rev{6}  \\
\rev{6}   & \rev{5}  & \rev{11} & \rev{14} & \rev{10} & \rev{8}  & \rev{60} & \rev{24} & \rev{15} & \rev{7}  & \rev{7}  \\
\rev{7}   & \rev{3}  & \rev{7}  & \rev{7}  & \rev{2}  & \rev{6}  & \rev{44} & \rev{9}  & \rev{11} & \rev{8}  & \rev{8}  \\
\rev{13}  & \rev{10} & \rev{13} & \rev{12} & \rev{7}  & \rev{12} & \rev{10} & \rev{17} & \rev{10} & \rev{9}  & \rev{9}  \\
\rev{10}  & \rev{15} & \rev{14} & \rev{13} & \rev{8} & \rev{11} & \rev{70} & \rev{21} & \rev{33} & \rev{10} & \rev{10} \\ \hline
\end{tabular}
}
\end{table}

\begin{table}[htbp]
  \caption{\rev{Predicted Rankings of the Top 10 Nodes in the Ground Truth of RD-Net3478.}}
  \label{tab:top10-3478}
  \centering
  \small
  \resizebox{\linewidth}{!}{%
\begin{tabular}{|ccccccccccc|}
\hline
\rev{DeepWalk} & \rev{GraphRNA} & \rev{TraNode2vec} & \rev{MGL2Rank} & \rev{HGT} & \rev{Dir-GNN} & \rev{GAT} & \rev{ENGINE} & \rev{\textnormal{GNN\textsubscript{LLMEmb}}} & \rev{GATv2} & \rev{PDFormer} \\ \hline
\rev{237} & \rev{3} & \rev{466} & \rev{1} & \rev{1} & \rev{96} & \rev{5} & \rev{184} & \rev{376} & \rev{12} & \rev{1} \\
\rev{443} & \rev{19} & \rev{41} & \rev{4} & \rev{6} & \rev{50} & \rev{34} & \rev{16} & \rev{1} & \rev{2} & \rev{4} \\
\rev{610} & \rev{4} & \rev{107} & \rev{2} & \rev{3} & \rev{2} & \rev{4} & \rev{50} & \rev{4} & \rev{6} & \rev{2} \\
\rev{3} & \rev{1} & \rev{217} & \rev{3} & \rev{2} & \rev{31} & \rev{3} & \rev{101} & \rev{7} & \rev{15} & \rev{3} \\
\rev{1} & \rev{86} & \rev{665} & \rev{19} & \rev{37} & \rev{8} & \rev{135} & \rev{312} & \rev{5} & \rev{54} & \rev{18} \\
\rev{612} & \rev{56} & \rev{18} & \rev{16} & \rev{22} & \rev{23} & \rev{81} & \rev{8} & \rev{35} & \rev{3} & \rev{15} \\
\rev{599} & \rev{11} & \rev{169} & \rev{5} & \rev{10} & \rev{117} & \rev{15} & \rev{88} & \rev{27} & \rev{19} & \rev{5} \\
\rev{409} & \rev{20} & \rev{46} & \rev{11} & \rev{15} & \rev{26} & \rev{31} & \rev{28} & \rev{15} & \rev{7} & \rev{10} \\
\rev{405} & \rev{15} & \rev{22} & \rev{9} & \rev{12} & \rev{82} & \rev{18} & \rev{12} & \rev{242} & \rev{4} & \rev{8} \\
\rev{187} & \rev{47} & \rev{49} & \rev{18} & \rev{30} & \rev{87} & \rev{62} & \rev{29} & \rev{8} & \rev{11} & \rev{19} \\ \hline
\rev{DR-HGNN} & \rev{SeHGNN} & \rev{Edge-GNN} & \rev{LambdaLoss} & \rev{RankFormer} & \rev{LightGBM} & \rev{IPageRank} & \rev{CRRank} & \rev{OVED-Rank} & \rev{HetGL2R} & \rev{Ground-Truth} \\ \hline
\rev{14} & \rev{3} & \rev{6} & \rev{18} & \rev{1} & \rev{3} & \rev{385} & \rev{4} & \rev{15} & \rev{2} & \rev{1} \\
\rev{2} & \rev{1} & \rev{1} & \rev{11} & \rev{2} & \rev{2} & \rev{445} & \rev{1} & \rev{4} & \rev{1} & \rev{2} \\
\rev{6} & \rev{7} & \rev{4} & \rev{1} & \rev{4} & \rev{1} & \rev{568} & \rev{12} & \rev{6} & \rev{3} & \rev{3} \\
\rev{15} & \rev{20} & \rev{9} & \rev{9} & \rev{7} & \rev{5} & \rev{3} & \rev{2} & \rev{2} & \rev{5} & \rev{4} \\
\rev{52} & \rev{5} & \rev{27} & \rev{3} & \rev{9} & \rev{4} & \rev{2} & \rev{5} & \rev{14} & \rev{9} & \rev{5} \\
\rev{3} & \rev{26} & \rev{3} & \rev{8} & \rev{5} & \rev{6} & \rev{310} & \rev{19} & \rev{11} & \rev{4} & \rev{6} \\
\rev{19} & \rev{66} & \rev{15} & \rev{46} & \rev{13} & \rev{14} & \rev{458} & \rev{16} & \rev{3} & \rev{10} & \rev{7} \\
\rev{7} & \rev{21} & \rev{7} & \rev{13} & \rev{12} & \rev{7} & \rev{494} & \rev{21} & \rev{20} & \rev{7} & \rev{8} \\
\rev{4} & \rev{191} & \rev{5} & \rev{37} & \rev{10} & \rev{12} & \rev{486} & \rev{34} & \rev{21} & \rev{6} & \rev{9} \\
\rev{10} & \rev{15} & \rev{10} & \rev{43} & \rev{17} & \rev{19} & \rev{247} & \rev{15} & \rev{36} & \rev{11} & \rev{10} \\ \hline
\end{tabular}
}
\end{table}

\rev{Tables \ref{tab:top10-110} to \ref{tab:top10-3478} present the predicted rankings of the ground-truth top-10 nodes across the three networks. The results indicate that HetGL2R achieves the highest ranking consistency, while PDFormer delivers competitive performance. Both models demonstrate robust ranking stability across different network configurations.}

\begin{figure}[ht]
    \centering
    \includegraphics[width=0.6\linewidth]{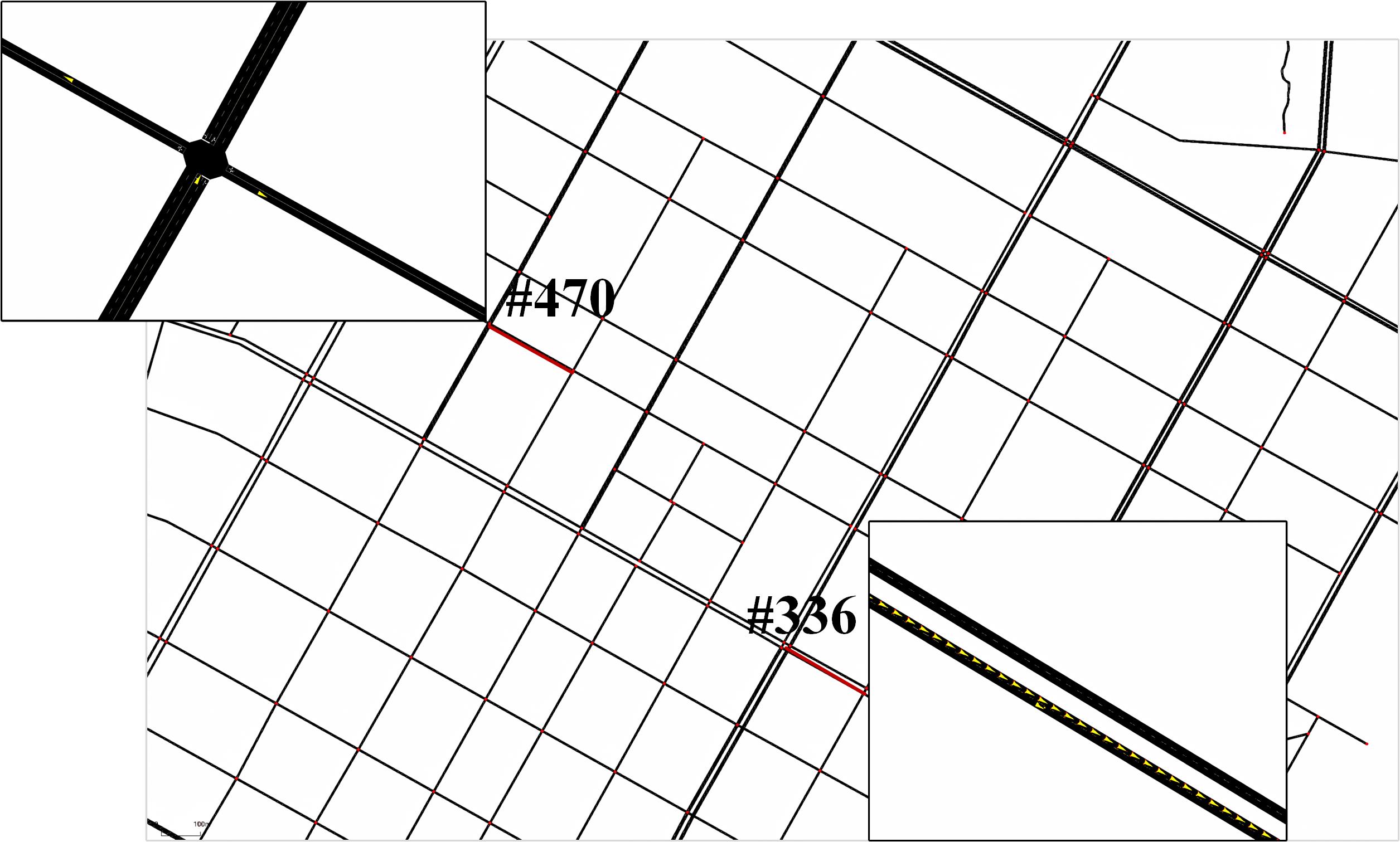}
    \caption{\revv{Case study on segment ranking in the SY-Net514 network. Two highly ranked road segments (\#470 and \#336) identified by HetGL2R are highlighted in red to illustrate their structural positions in the network.}}
    \label{fig:g4}
\end{figure}

\begin{table}[htbp]
\caption{\rev{Runtime Usage of HetGL2R and Baseline Methods on Three Networks.}}
\label{tab:time}
\centering
\tablesize
\begin{tabular}{c|ccc}
\toprule\hline
\multirow{2}{*}{\rev{Method}} & \multicolumn{3}{c}{\rev{Time(s)}}       \\ \cline{2-4} 
                        & \rev{SY-Net110} & \rev{SY-Net514} & \rev{RD-Net3478}  \\ \hline
\rev{DeepWalk}                & \rev{7.12}      & \rev{41.93}     & \rev{299.71}        \\
\rev{GraphRNA}                & \rev{17.83}     & \rev{149.82}    & \rev{326.18}       \\
\rev{TraNode2vec}             & \rev{8.15}      & \rev{56.12}     & \rev{378.29}      \\
\rev{MGL2Rank}                & \rev{13.83}     & \rev{123.63}    & \rev{302.25}       \\
\rev{HGT}                     & \rev{17.54}     & \rev{104.89}    & \rev{190.25}      \\
\rev{Dir-GNN}                 & \rev{16.77}     & \rev{108.86}    & \rev{211.84}       \\
\rev{GAT}                     & \rev{12.03}     & \rev{124.19}    & \rev{311.47}       \\
\rev{ENGINE}                  & \rev{15.18}     & \rev{217.56}    & \rev{882.36}       \\
\rev{\textnormal{GNN\textsubscript{LLMEmb}}}               & \rev{28.40}     & \rev{300.18}    & \rev{680.40}      \\
\rev{GATv2}                   & \rev{15.42}     & \rev{147.18}    & \rev{412.34}        \\
\rev{PDFormer}                & \rev{125.21}    & \rev{321.43}    & \rev{1090.57}       \\
\rev{DR-HGNN}                 & \rev{152.33}    & \rev{473.79}    & \rev{1544.77}       \\
\rev{SeHGNN}                  & \rev{31.61}     & \rev{115.87}    & \rev{341.36}       \\
\rev{Edge-GNN}                & \rev{128.89}    & \rev{381.57}    & \rev{1130.66}       \\
\rev{LambdaLoss}              & \rev{9.31}      & \rev{13.09}     & \rev{122.63}       \\
\rev{RankFormer}              & \rev{11.28}     & \rev{16.96}     & \rev{163.25}       \\
\rev{LightGBM}                & \rev{2.56}      & \rev{4.13}      & \rev{13.27}       \\
\rev{HetGL2R}                 & \rev{16.27}     & \rev{126.93}    & \rev{314.92}      \\ \hline\bottomrule
\end{tabular}
\end{table}

\rev{Figure \ref{fig:g4} illustrates two representative road segments during ranking evaluation on the SY-Net514 network. In the rankings generated by HetGL2R, segment \#470 has a relatively low traffic volume (215 vehicles/hour) but is ranked 15th, higher than segment \#336, which carries a larger traffic volume (588 vehicles/hour, ranked 22nd). Further analysis shows that segment \#470 is located near the destination of a certain OD pair, with upstream nodes carrying substantial traffic flow. A failure at segment \#470 could block upstream traffic, causing congestion in the upstream region and potentially impacting the broader network performance. This example illustrates that HetGL2R accounts for upstream–downstream dependencies along traffic paths, beyond node attributes or local topological features.}

\subsection{\rev{Computational Efficiency and Scalability Analysis}}
Table \ref{tab:time} summarizes the per-epoch running time of HetGL2R on three networks with increasing scales. Compared with computationally lightweight methods (e.g., DeepWalk and TraNode2vec), specialized ranking models (e.g., LambdaLoss and RankFormer), and the ensemble model LightGBM, HetGL2R incurs higher training cost due to its more expressive architecture. However, when compared with other graph neural network–based approaches (e.g., PDFormer, DR-HGNN, and Edge-GNN), HetGL2R exhibits comparable computational efficiency. This efficiency can be attributed to its random-walk-based sampling strategy, which constrains computational complexity to scale linearly with walk length and sampling count. In contrast, models relying on global attention mechanisms or deep hierarchical message passing often incur quadratic or higher-order complexity with respect to the number of nodes. The primary scalability bottlenecks of HetGL2R stem from the Transformer encoder’s self-attention operations over sampled walk sequences and the storage and I/O overhead associated with full-graph walk sampling. As a result, while HetGL2R may not match the efficiency of lightweight models that do not exploit graph structure, it is expected to scale more favorably than most GNN-based methods on large graphs. Overall, by trading sampling for computation, HetGL2R achieves a practical balance between modeling expressiveness and computational feasibility.

\begin{table}[t]
  \caption{\rev{Ablation Study on Ranking Performance under four Views.}}
  \label{tab:ablation_modules}
  \centering
  \small
  \resizebox{\linewidth}{!}{%
  \begin{tabular}{c|c c c c |c c c c |c c c c}
    \toprule
    \hline
    \multirow{2}{*}{\rev{module}} &
      \multicolumn{4}{c!{\vrule width 0.6pt}}{\rev{SY-Net110}} &
      \multicolumn{4}{c!{\vrule width 0.6pt}}{\rev{SY-Net514}} &
      \multicolumn{4}{c}{\rev{RD-Net3478}} \\
      \cline{2-5}\cline{6-9}\cline{10-13}
    & \rev{EMD} & \rev{NDCG@K} & \rev{Diff} & \rev{$\tau$}
      & \rev{EMD} & \rev{NDCG@K} & \rev{Diff} & \rev{$\tau$}
      & \rev{EMD} & \rev{NDCG@K} & \rev{Diff} & \rev{$\tau$} \\
    \cline{1-5}\cline{6-9}\cline{10-13}
    \rev{w/o AG}                & \rev{0.2198} & \rev{0.9071} & \rev{0.2811} & \rev{0.7284}
                                & \rev{0.1224} & \rev{0.8760} & \rev{0.3053} & \rev{0.7919}
                                & \rev{0.3275} & \rev{0.7774} & \rev{0.3806} & \rev{0.7117} \\
    \rev{w/o transformer(GRU)}  & \rev{0.1564} & \rev{0.7882} & \rev{0.3813} & \rev{0.6103}
                                & \rev{0.1698} & \rev{0.7714} & \rev{0.3761} & \rev{0.7866}
                                & \rev{0.4048} & \rev{0.6569} & \rev{0.4579} & \rev{0.5906} \\
    \rev{w/o transformer(LSTM)} & \rev{0.2700} & \rev{0.7646} & \rev{0.4021} & \rev{0.6829}
                                & \rev{0.2695} & \rev{0.7693} & \rev{0.3992} & \rev{0.7703}
                                & \rev{0.3746} & \rev{0.7393} & \rev{0.3930} & \rev{0.6634} \\
    \rev{W/o AMIL(Mean)}        & \rev{\textbf{0.0637}} & \rev{0.9116} & \rev{0.2536} & \rev{0.7335}
                                & \rev{\textbf{0.1221}} & \rev{0.9380} & \rev{\textbf{0.2404}} & \rev{0.8638}
                                & \rev{0.3882} & \rev{0.8236} & \rev{0.3402} & \rev{0.7302} \\
    \rev{W/o AMIL(DeepSets)}    & \rev{0.0788} & \rev{0.9215} & \rev{0.2533} & \rev{0.7432}
                                & \rev{0.1474} & \rev{0.9252} & \rev{0.2645} & \rev{0.8591}
                                & \rev{0.3726} & \rev{0.8344} & \rev{0.3461} & \rev{0.7574} \\
    \rev{W/o AMIL(SetTrans)}    & \rev{0.0705} & \rev{0.9246} & \rev{0.2834} & \rev{0.7724}
                                & \rev{0.1421} & \rev{0.9458} & \rev{0.2648} & \rev{0.8741}
                                & \rev{0.3940} & \rev{0.8468} & \rev{0.3612} & \rev{0.7759} \\
    \rev{W/o KL(ListNet)}       & \rev{0.0761} & \rev{0.9190} & \rev{0.2814} & \rev{0.7485}
                                & \rev{0.2607} & \rev{0.9054} & \rev{0.2744} & \rev{0.8478}
                                & \rev{0.3512} & \rev{0.8034} & \rev{0.3526} & \rev{0.7462} \\
    \rev{W/o KL(ListMLE)}       & \rev{0.0939} & \rev{0.9108} & \rev{0.2830} & \rev{0.7341}
                                & \rev{0.2340} & \rev{0.8483} & \rev{0.3655} & \rev{0.7208}
                                & \rev{0.3670} & \rev{0.7467} & \rev{0.4018} & \rev{0.7028} \\
    \hline
    \rev{HetGL2R}               & \rev{0.0699} & \textbf{\rev{0.9459}} & \textbf{\rev{0.2483}} & \textbf{\rev{0.7934}}
                                & \rev{0.1318} & \textbf{\rev{0.9471}} & \rev{0.2478} & \rev{\textbf{0.8898}}
                                & \textbf{\rev{0.3208}} & \textbf{\rev{0.8502}} & \textbf{\rev{0.3318}} & \textbf{\rev{0.7918}} \\
    \hline
    \bottomrule
  \end{tabular}%
  }
\end{table}

\subsection{\rev{Ablation Study}}
\rev{We conducted ablation studies on HetGWalk, the Transformer encoder, the AMIL aggregation function, and the KL-divergence loss, with results shown in Table \ref{tab:ablation_modules}.}

Ablation Study on HetGWalk. We evaluated the impact of joint random walks on the Trip Graph (TG) and Attribute-guided Graph (AG) by comparing the full model with a variant excluding AG (w/o AG). The w/o AG variant consistently shows substantial performance degradation across all networks and metrics, indicating that joint random walks are crucial for accurate and robust ranking. Using TG alone captures OD-induced functional dependencies along the graph structure, but does not explicitly model functional similarity, which can bias sampling toward high-centrality nodes and underexplore latent functional clusters. By enabling transitions between semantically similar nodes, AG enriches contextual diversity. Consequently, the joint TG–AG random walk integrates OD-induced functional dependence and semantic similarity, producing more informative sequences for robust segment representation.

Ablation on Transformer Encoder. We evaluated the role of the Transformer encoder by replacing it with GRU \citep{gru} and LSTM \citep{lstm}, denoted as w/o Transformer (GRU) and w/o Transformer (LSTM), respectively. The w/o Transformer variants consistently show inferior performance across all datasets and metrics, indicating that the Transformer encoder plays an important role in HetGL2R. RNN-based encoders rely on sequential propagation, which makes modeling long-range dependencies less effective in random-walk sequences. In contrast, the Transformer’s global self-attention mechanism enables direct interactions between arbitrary positions in a sequence, supporting more effective modeling of path-level dependencies and contributing to more robust ranking performance.

Ablation Study on AMIL Aggregation. We examined the impact of different context aggregation strategies by replacing AMIL with Mean, DeepSets \citep{deepsets}, and SetTrans \citep{SetTrans}, denoted as w/o AMIL (Mean), w/o AMIL (DeepSets), and w/o AMIL (SetTrans), respectively. The variants without AMIL consistently show inferior performance across datasets and metrics, indicating the effectiveness of attention-based aggregation in HetGL2R. Mean aggregation treats all contexts equally and may be less effective at distinguishing critical contexts, while DeepSets aggregates instances independently and does not explicitly model inter-instance interactions. SetTrans, although expressive, may place excessive emphasis on fine-grained variations, making it more sensitive to noise. In contrast, AMIL employs a lightweight attention mechanism to selectively emphasize informative contexts while suppressing irrelevant ones, achieving a better balance between robustness and representation quality.

Ablation Study on KL-Divergence Loss. We examined the effect of different listwise ranking losses by replacing the KL-divergence loss with the cross-entropy loss used in ListNet \citep{listnet} and the likelihood-based loss used in ListMLE \citep{mle}, denoted as w/o KL (ListNet) and w/o KL (ListMLE), respectively. The variants without KL divergence consistently show inferior performance across datasets and metrics, indicating the effectiveness of KL divergence for this task. This performance gap can be attributed to differences in modeling assumptions among listwise losses. ListMLE relies on stepwise ranking generation, which may accumulate errors when ranking supervision is uncertain, while ListNet focuses primarily on the Top-one probability and does not fully utilize ranking information from the remaining positions. In contrast, KL divergence performs a soft alignment between predicted and ground-truth importance distributions, emphasizing top-ranked segments while leveraging supervision from all positions, without relying on stepwise generation assumptions.

\begin{figure}[t]
    \centering
    \includegraphics[width=1\linewidth]{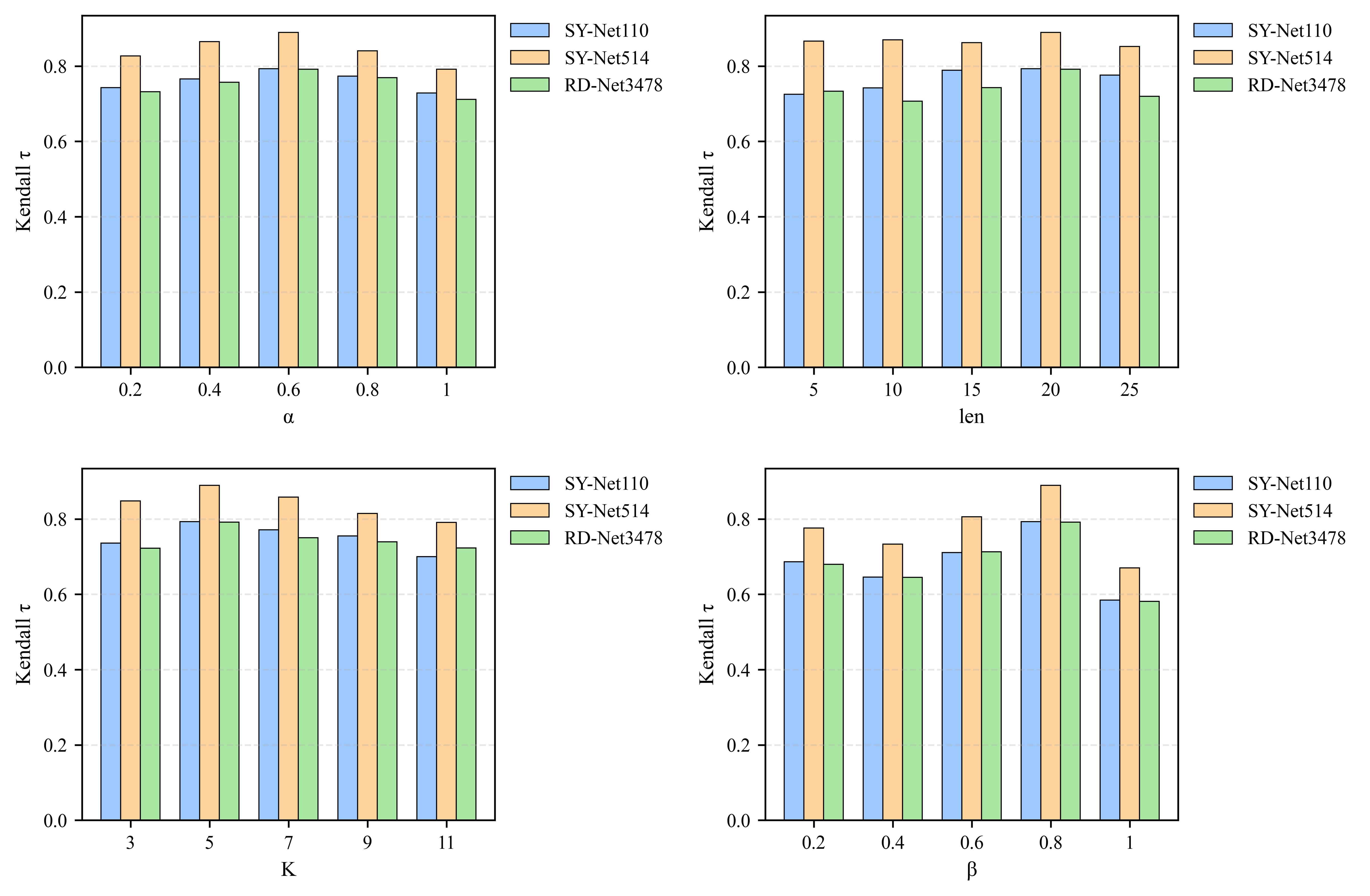}
    \caption{\revv{Parameter sensitivity analysis on three datasets. Each sub-plot shows the effect of a specific hyperparameter on model performance. The horizontal axis corresponds to the variation of the mixing coefficient $\alpha$, sampling sequence length $len$, list length $K$, and depth-first exploration probability $\beta$, respectively, while the vertical axis represents the Kendall’s $\tau$ score.}}
    \label{fig:lmd}
\end{figure}

Overall, these ablation results suggest that each core component in the HetGL2R framework contributes meaningfully to the final performance. Their combination enables a balanced trade-off between modeling expressiveness and robustness for critical road segment identification.

\subsection{Sensitivity analysis}
We conducted a sensitivity analysis on key hyperparameters, including the walk bias factor $\alpha$, the sampling sequence length $len$, the list length $K$, and the depth-first exploration probability $\beta$, with results summarized in Figure \ref{fig:lmd}. When $\alpha$ is set to 0.6, the model achieves the highest Kendall’s $\tau$, suggesting a favorable trade-off between topological and attribute-related information. The optimal sampling length $len$ is 20, as shorter sequences provide limited contextual coverage, while longer sequences may introduce redundant or noisy information that degrades performance. Similarly, a list length of $K = 5$ yields the best performance, as shorter lists offer insufficient ranking context, whereas longer lists increase task complexity and dilute effective ranking signals. Finally, setting $\beta$ to 0.8 results in the highest Kendall’s $\tau$, indicating that a moderate preference for depth-first exploration is beneficial. Smaller $\beta$ values underexploit multi-hop context, while $\beta =1.0$ shifts the walk toward a breadth-first–dominated exploration pattern, weakening deep relational signals and leading to noticeable performance degradation. Overall, these results demonstrate that HetGL2R is sensitive to key hyperparameters and provide empirical guidance for practical parameter selection.

\subsection{\revvv{Discussion}}
\subsubsection{\revvv{Analysis of Model Performance}}
The observed performance differences across baselines can be attributed to differences in their modeling assumptions and inductive biases, particularly in how structural cues, long-range dependencies, and path-based spatial influence are represented. Topology-dependent methods such as DeepWalk and IPageRank primarily rely on local connectivity patterns and centrality-related signals. These assumptions are often reasonable in structured urban networks, where network topology and OD flow are closely correlated. However, when this coupling weakens, their effectiveness degrades. This helps explain the pronounced performance drop observed in RD-Net3478, where network connectivity is generated independently of OD flow patterns, and highly connected or centrally located segments do not necessarily correspond to traffic corridors carrying substantial OD flow. Message-passing-based GNNs, including GAT and Dir-GNN, are further constrained by limited receptive fields and locality of aggregation. Even with multiple layers, information propagation is largely restricted to physical adjacency, making it challenging to capture long-range, path-dependent influence patterns induced by OD flows. More advanced heterogeneous models such as HGT and SeHGNN exhibit improved robustness on structured networks by incorporating type-aware aggregation and extended meta-paths. However, their reliance on message passing limits their ability to exploit sequential path semantics, such as node order and directional influence along OD paths, which are important for modeling traffic directionality and downstream effects. Notably, PDFormer achieves consistently strong performance across all three datasets. As a traffic-oriented model, it leverages semantic-mask-guided attention to capture long-range dependencies between geographically distant yet functionally related nodes and incorporates propagation-delay-aware mechanisms. While conceptually aligned with HetGL2R in emphasizing functional and long-range dependencies, PDFormer models such dependencies implicitly through masked attention over spatiotemporal views, which may be limited by error accumulation in similarity measurement (e.g., DTW-based semantic masking) over long distances due to signal attenuation and local noise interference. In contrast, HetGL2R explicitly represents OD flows, paths, and segment-level interactions, enabling a more fine-grained and explicit representation of traffic-induced influence propagation. Graph-based large language model approaches, such as ENGINE and \textnormal{GNN\textsubscript{LLMEmb}}, exhibit limited effectiveness in this task. Although they provide rich semantic priors, they are designed for general graph structures and textual relational reasoning. Without explicit mechanisms to encode OD flows, path dependencies, or traffic-specific propagation dynamics, these models struggle to align their representations with the underlying mechanisms governing spatial influence in transportation networks. Overall, these observations suggest that the superior performance of HetGL2R is not solely due to the inclusion of OD information, but also to its sequence-based representation of spatial influence along traffic paths, which better aligns with the functional mechanisms underlying road-segment criticality.

\subsubsection{\revvv{Theoretical Implications}}
\revvv{Building on the analysis above, we distill the broader theoretical insights. First, our findings provide empirical evidence that functional structure—induced by OD flows and paths—plays a more influential role than physical topology in modeling long-range spatial dependencies in road networks. This observation challenges the conventional reliance on purely topological measures and suggests that OD-aware modeling is important for understanding traffic-induced spatial influence. Second, by reframing road segment ranking as a learning-to-rank problem on heterogeneous graphs, our work introduces an alternative methodological lens for transportation network analysis. This formulation enables the explicit integration of OD flows and path structures—elements that are difficult to incorporate into standard graph neural network architectures—within a unified, task-driven learning framework. Third, our work offers an alternative perspective on the notion of "importance" in networked systems. Unlike static centrality measures that treat importance as an inherent node property, our framework conceptualizes importance as context-dependent, shaped by demand patterns and usage paths.}

\subsubsection{\revvv{Practical Implications}}
\revvv{From a practical perspective, the resulting road segment rankings are intended to serve as a decision-support signal, providing an initial prioritization of potentially critical segments that can be combined with domain-specific operational objectives and policy constraints. Possible application scenarios include:}

\begin{itemize}
    \item \revvv{Resilience assessment and monitoring. Transportation agencies can use the rankings to help identify segments whose failure may induce disproportionate network-wide disruption, supporting targeted monitoring and proactive reinforcement of high-risk corridors.}

    \item \revvv{Emergency response planning. By highlighting segments that are critical to high-demand OD pairs, the rankings can inform the design of contingency routing strategies that reduce reliance on vulnerable corridors during emergencies such as natural disasters or major incidents.}

    \item \revvv{Infrastructure investment prioritization. Under limited maintenance and upgrade budgets, the rankings can support the prioritization of road segments with higher potential systemic impact, complementing existing engineering and policy-based decision criteria.}

    \item \revvv{Transferability to other infrastructure networks. Although demonstrated on road networks, the core modeling principle—capturing functional dependencies induced by OD flows—is conceptually transferable to other infrastructure systems, such as power grids (electricity demand between substations) or communication networks (data flows between routers).}
\end{itemize}

\revvv{Together, these examples illustrate how the proposed framework can support practical decision-making by providing structured, system-level insights, while final actions remain guided by domain expertise and operational constraints.}

\section{Conclusion and Future Work}
\rev{This paper proposes HetGL2R, a heterogeneous graph learning framework for traffic network ranking. The framework employs heterogeneous random walks and Transformer architecture to jointly model the functional relationships among OD flows, paths, and road segments along with their attribute similarities. By integrating functional and structural information, it effectively captures long-range spatial dependencies, generating context-aware road segment embeddings. These embeddings are subsequently processed by a listwise ranking module to produce the final importance ranking. Extensive evaluations on three simulated networks validate the framework’s superior accuracy for ranking road networks.}

\revv{While the experimental evaluation is conducted on simulation-based networks, this choice reflects practical constraints in real-world traffic analysis, where large-scale controlled disruptions involve substantial cost, time, and institutional barriers. Simulation therefore provides a feasible environment for quantifying system-level criticality. Looking ahead, future work will focus on transferring the proposed framework to real-world settings by improving robustness to noisy or incomplete OD demand estimates, incorporating uncertainty-aware modeling, and validating the approach on large-scale operational networks as suitable data become available.}


\section*{Statement}
During the preparation of this work the author(s) used DeepSeek and GPT4 in order to enhance the linguistic fluency of the manuscript. After using this tool/service, the author(s) reviewed and edited the content as needed and take(s) full responsibility for the content of the publication.


\bibliographystyle{elsarticle-harv} 
\bibliography{reference}

@article{rf1,
  author={Xu, Ming and Wu, Jianping and Liu, Mengqi and Xiao, Yunpeng and Wang, Haohan and Hu, Dongmei},
  title={Discovery of critical nodes in road networks through mining from vehicle trajectories},
  journal={IEEE Transactions on Intelligent Transportation Systems},
  volume={20},
  year={2018},
  pages={583--593}
}

@inproceedings{rf2,
  author={P. A. Lopez and M. Behrisch and L. Bieker-Walz and J. Erdmann and Y.-P. Flötteröd and R. Hilbrich and L. Lücken and J. Rummel and P. Wagner and E. Wießner},
  title={Microscopic traffic simulation using SUMO},
  booktitle={Proceedings of the 21st International Conference on Intelligent Transportation Systems (ITSC)},
  year={2018},
  pages={2575--2582}
}

@article{rf3,
  author={Agryzkov, Taras and Oliver, Jose L and Tortosa, Leandro and Vicent, Jose F},
  title={An algorithm for ranking the nodes of an urban network based on the concept of PageRank vector},
  journal={Applied Mathematics and Computation},
  volume={219},
  year={2012},
  pages={2186--2193}
}

@article{rf4,
  author  = {Fan, Changjun and Zeng, Li and Sun, Yizhou and Liu, Yang-Yu},
  title   = {Finding key players in complex networks through deep reinforcement learning},
  journal = {Nature machine intelligence},
  volume  = {2},
  year    = {2020},
  pages   = {317--324}
}

@inproceedings{rf5,
  author={W. Hamilton and Z. Ying and J. Leskovec},
  title={Inductive representation learning on large graphs},
  booktitle={Proceedings of the 30th Conference on Neural Information Processing Systems (NeurIPS)},
  location={Long Beach, CA, USA},
  volume={30},
  year={2017},
  pages={1024--1034}
}

@article{rf6,
  author  = {Tan, Xuwei and Zhou, Yangming and Zhou, MengChu and Fu, Zhang-Hua},
  title   = {Learning to detect critical nodes in sparse graphs via feature importance awareness},
  journal = {IEEE Transactions on Automation Science and Engineering},
  year    = {2024}
}

@article{rf7,
  author  = {Huang, Xinlong and Chen, Jian and Cai, Ming and Wang, Wei and Hu, Xiping},
  title   = {Traffic node importance evaluation based on clustering in represented transportation networks},
  journal = {IEEE Transactions on Intelligent Transportation Systems},
  volume  = {23},
  year    = {2022},
  pages   = {16622--16631}
}

@article{rf8,
  author  = {Xu, Ming and Zhang, Jing},
  title   = {MGL2Rank: Learning to rank the importance of nodes in road networks based on multi-graph fusion},
  journal = {Information Sciences},
  volume  = {667},
  year    = {2024},
  pages   = {120472}
}

@inproceedings{rf9,
  author  = {Abdollahpouri, Himan and Burke, Robin and Mobasher, Bamshad},
  title   = {Controlling popularity bias in learning-to-rank recommendation},
  booktitle = {Proceedings of the eleventh ACM conference on recommender systems},
  year    = {2017},
  pages   = {42--46}
}

@article{rf10,
  author  = {Song, Qiang and Liu, Anqi and Yang, Steve Y},
  title   = {Stock portfolio selection using learning-to-rank algorithms with news sentiment},
  journal = {Neurocomputing},
  volume  = {264},
  year    = {2017},
  pages   = {20--28}
}

@inproceedings{rf11,
  author  = {Tan, Zhiyu and Nie, Xuecheng and Qian, Qi and Li, Nan and Li, Hao},
  title   = {Learning to rank proposals for object detection},
  booktitle = {Proceedings of the IEEE/CVF International Conference on Computer Vision},
  year    = {2019},
  pages   = {8273--8281}
}

@article{rf12,
  author  = {Li, Ping and Wu, Qiang and Burges, Christopher},
  title   = {Mcrank: Learning to rank using multiple classification and gradient boosting},
  journal = {Advances in neural information processing systems},
  volume  = {20},
  year    = {2007}
}

@inproceedings{rf13,
  author  = {Burges, Chris and Shaked, Tal and Renshaw, Erin and Lazier, Ari and Deeds, Matt and Hamilton, Nicole and Hullender, Greg},
  title   = {Learning to rank using gradient descent},
  booktitle = {Proceedings of the 22nd international conference on Machine learning},
  year    = {2005},
  pages   = {89--96}
}

@article{rf14,
  author  = {Maurya, Sunil Kumar and Liu, Xin and Murata, Tsuyoshi},
  title   = {Graph neural networks for fast node ranking approximation},
  journal = {ACM Transactions on Knowledge Discovery from Data (TKDD)},
  volume  = {15},
  year    = {2021},
  pages   = {1--32}
}

@inproceedings{rf15,
  author  = {Xia, Fen and Liu, Tie-Yan and Wang, Jue and Zhang, Wensheng and Li, Hang},
  title   = {Listwise approach to learning to rank: theory and algorithm},
  booktitle = {Proceedings of the 25th international conference on Machine learning},
  year    = {2008},
  pages   = {1192--1199}
}

@inproceedings{rf16,
  author  = {Lan, Yanyan and Liu, Tie-Yan and Ma, Zhiming and Li, Hang},
  title   = {Generalization analysis of listwise learning-to-rank algorithms},
  booktitle = {Proceedings of the 26th Annual International Conference on Machine Learning},
  year    = {2009},
  pages   = {577--584}
}

@article{rf17,
  author  = {Wu, Ou and You, Qiang and Mao, Xue and Xia, Fen and Yuan, Fei and Hu, Weiming},
  title   = {Listwise learning to rank by exploring structure of objects},
  journal = {IEEE Transactions on Knowledge and Data Engineering},
  volume  = {28},
  year    = {2016},
  pages   = {1934--1939}
}

@article{rf18,
  author  = {Pobrotyn, Przemysław and Bartczak, Tomasz and Synowiec, Mikołaj and Białobrzeski, Radosław and Bojar, Jarosław},
  title   = {Context-aware learning to rank with self-attention},
  journal = {arXiv preprint arXiv:2005.10084},
  year    = {2020},
  doi={10.48550/arXiv.2005.10084}
}

@inproceedings{rf19,
  author  = {Ai, Qingyao and Bi, Keping and Guo, Jiafeng and Croft, W Bruce},
  title   = {Learning a deep listwise context model for ranking refinement},
  booktitle = {The 41st international ACM SIGIR conference on research \& development in information retrieval},
  year    = {2018},
  pages   = {135--144}
}

@inproceedings{rf20,
  author  = {Zhu, Xiaofeng and Klabjan, Diego},
  title   = {Listwise learning to rank by exploring unique ratings},
  booktitle = {Proceedings of the 13th international conference on web search and data mining},
  year    = {2020},
  pages   = {798--806}
}

@inproceedings{rf21,
  author  = {Singh, Sonali and Farfade, Sachin and Comar, Prakash},
  title   = {DiAL: Diversity aware listwise ranking for query auto-complete},
  booktitle = {Proceedings of the 2024 Conference on Empirical Methods in Natural Language Processing: Industry Track},
  year    = {2024},
  pages   = {1152--1162}
}

@article{rf22,
  author  = {Yang, Sihui and Bi, Keping and Cui, Wanqing and Guo, Jiafeng and Cheng, Xueqi},
  title   = {LINKAGE: Listwise Ranking among Varied-Quality References for Non-Factoid QA Evaluation via LLMs},
  journal = {arXiv preprint arXiv:2409.14744},
  year    = {2024},
  doi={10.48550/arXiv.2409.14744}
}

@inproceedings{rf23,
  author={Thonet, Thibaut and Cinar, Yagmur Gizem and Gaussier, Eric and Li, Minghan and Renders, Jean-Michel},
  title={Listwise learning to rank based on approximate rank indicators},
  booktitle={Proceedings of the AAAI Conference on Artificial Intelligence},
  volume={36},
  year={2022},
  pages={8494--8502},
}

@inproceedings{rf24,
  author  = {Bai, Aijun and Jagerman, Rolf and Qin, Zhen and Yan, Le and Kar, Pratyush and Lin, Bing-Rong and Wang, Xuanhui and Bendersky, Michael and Najork, Marc},
  title   = {Regression compatible listwise objectives for calibrated ranking with binary relevance},
  booktitle = {Proceedings of the 32nd ACM International Conference on Information and Knowledge Management},
  year    = {2023},
  pages   = {4502--4508}
}

@article{rf25,
  author  = {Shi, Chuan and Hu, Binbin and Zhao, Wayne Xin and Yu, Philip S},
  title   = {Heterogeneous information network embedding for recommendation},
  journal = {IEEE transactions on knowledge and data engineering},
  volume  = {31},
  year    = {2018},
  pages   = {357--370}
}

@inproceedings{rf26,
  author  = {Dong, Yuxiao and Tang, Jie and Wu, Sen and Tian, Jilei and Chawla, Nitesh V and Rao, Jinghai and Cao, Huanhuan},
  title   = {Link prediction and recommendation across heterogeneous social networks},
  booktitle = {2012 IEEE 12th International Conference on Data Mining},
  year    = {2012},
  pages   = {181--190}
}

@inproceedings{rf27,
  author={Li, Xiaoxue and Shang, Yanmin and Cao, Yanan and Li, Yangxi and Tan, Jianlong and Liu, Yanbing},
  title={Type-aware anchor link prediction across heterogeneous networks based on graph attention network},
  booktitle={Proceedings of the AAAI Conference on Artificial Intelligence},
  volume={34},
  year={2020},
  pages={147--155}
}

@inproceedings{rf28,
  author  = {X. Wang and H. Ji and C. Shi and B. Wang and Y. Ye and P. Cui and P. S. Yu},
  title   = {Heterogeneous graph attention network},
  booktitle = {Proc. World Wide Web Conf.},
  year    = {2019},
  pages   = {2022--2032}
}

@inproceedings{rf29,
  author  = {Fu, Xinyu and Zhang, Jiani and Meng, Ziqiao and King, Irwin},
  title   = {Magnn: Metapath aggregated graph neural network for heterogeneous graph embedding},
  booktitle = {Proceedings of the web conference 2020},
  year    = {2020},
  pages   = {2331--2341}
}

@article{rf30,
  author  = {Chang, Yaomin and Chen, Chuan and Hu, Weibo and Zheng, Zibin and Zhou, Xiaocong and Chen, Shouzhi},
  title   = {Megnn: Meta-path extracted graph neural network for heterogeneous graph representation learning},
  journal = {Knowledge-Based Systems},
  volume  = {235},
  year    = {2022},
  pages   = {107611}
}

@inproceedings{rf31,
  author  = {Zhang, Chuxu and Song, Dongjin and Huang, Chao and Swami, Ananthram and Chawla, Nitesh V},
  title   = {Heterogeneous graph neural network},
  booktitle = {Proceedings of the 25th ACM SIGKDD international conference on knowledge discovery \& data mining},
  year    = {2019},
  pages   = {793--803}
}

@inproceedings{rf32,
  author={Hong, Huiting and Guo, Hantao and Lin, Yucheng and Yang, Xiaoqing and Li, Zang and Ye, Jieping},
  title={An attention-based graph neural network for heterogeneous structural learning},
  booktitle={Proceedings of the AAAI conference on artificial intelligence},
  volume={34},
  year={2020},
  pages={4132--4139}
}

@inproceedings{rf34,
  author={Samy, Ahmed E and Giaretta, Lodovico and Kefato, Zekarias T and Girdzijauskas, {\v{S}}ar{\=u}nas},
  title={Schemawalk: Schema aware random walks for heterogeneous graph embedding},
  booktitle={Companion Proceedings of the Web Conference 2022},
  year={2022},
  pages={1157--1166}
}

@inproceedings{rf35,
  author  = {Hussein, Rana and Yang, Dingqi and Cudré-Mauroux, Philippe},
  title   = {Are meta-paths necessary? Revisiting heterogeneous graph embeddings},
  booktitle = {Proceedings of the 27th ACM international conference on information and knowledge management},
  year    = {2018},
  pages   = {437--446}
}

@article{rf36,
  author  = {Guan, Mengya and Cai, Xinjun and Shang, Jiaxing and Hao, Fei and Liu, Dajiang and Jiao, Xianlong and Ni, Wancheng},
  title   = {Hmsg: Heterogeneous graph neural network based on metapath subgraph learning},
  journal = {Knowledge-Based Systems},
  volume  = {279},
  year    = {2023},
  pages   = {110930}
}

@article{rf37,
  author  = {Veličković, Petar and Fedus, William and Hamilton, William L and Liò, Pietro and Bengio, Yoshua and Hjelm, R Devon},
  title   = {Deep graph infomax},
  journal = {arXiv preprint arXiv:1809.10341},
  year    = {2018}
}

@inproceedings{rf38,
  author  = {He, Yu and Song, Yangqiu and Li, Jianxin and Ji, Cheng and Peng, Jian and Peng, Hao},
  title   = {Hetespaceywalk: A heterogeneous spacey random walk for heterogeneous information network embedding},
  booktitle = {Proceedings of the 28th ACM international conference on information and knowledge management},
  year    = {2019},
  pages   = {639--648}
}

@inproceedings{rf39,
  author  = {Perozzi, Bryan and Al-Rfou, Rami and Skiena, Steven},
  title   = {Deepwalk: Online learning of social representations},
  booktitle = {Proceedings of the 20th ACM SIGKDD international conference on Knowledge discovery and data mining},
  year    = {2014},
  pages   = {701--710}
}

@inproceedings{rf40,
  author  = {Huang, Xiao and Song, Qingquan and Li, Yuening and Hu, Xia},
  title   = {Graph recurrent networks with attributed random walks},
  booktitle = {Proceedings of the 25th ACM SIGKDD international conference on knowledge discovery \& data mining},
  year    = {2019},
  pages   = {732--740}
}

@article{rf41,
  author  = {Ruan, Yirun and Liu, Sizheng and Tang, Jun and Guo, Yanming and Yu, Tianyuan},
  title   = {GLC: A dual-perspective approach for identifying influential nodes in complex networks},
  journal = {Expert Systems with Applications},
  volume  = {268},
  year    = {2025},
  pages   = {126292}
}

@article{rf42,
  author  = {Ullah, Aman and Shao, Junming and Yang, Qinli and Khan, Nasrullah and Bernard, Cobbinah M. and Kumar, Rajesh},
  title   = {LSS: A locality-based structure system to evaluate the spreader’s importance in social complex networks},
  journal = {Expert Systems with Applications},
  volume  = {228},
  year    = {2023},
  pages   = {120326}
}

@article{rf43,
  author  = {Moradi, Mahnaz and Moradi, Parham and Faroughi, Azadeh and Jalili, Mahdi},
  title   = {Two-level attention mechanism with contrastive learning for heterogeneous graph representation learning},
  journal = {Expert Systems with Applications},
  volume  = {273},
  year    = {2025},
  pages   = {126751}
}

@article{rf44,
  author  = {Aman Ullah and Yahui Meng and J.~F.~F. Mendes},
  title   = {{OVED-Rank: A ranking scheme to evaluate complex network spreaders’ influence through the concept of effective distance and orbital velocity}},
  journal = {Information Processing \& Management},
  volume  = {62},
  year    = {2025},
  pages   = {104201}
}

@article{rf45,
  author  = {Yinghao Xiong and Xinhui Tu and Weizhong Zhao},
  title   = {{AFR-Rank: An effective and highly efficient LLM-based listwise reranking framework via filtering noise documents}},
  journal = {Information Processing \& Management},
  volume  = {62},
  year    = {2025},
  pages   = {104232}
}

@article{rf47,
  author  = {Yaling Xun and Yujia Wang and Jifu Zhang and Haifeng Yang and Jianghui Cai},
  title   = {{Higher-order embedded learning for heterogeneous information networks and adaptive POI recommendation}},
  journal = {Information Processing \& Management},
  year    = {2024},
  pages   = {103763}
}

@inproceedings{rf48,
    author = {Hu, Ziniu and Dong, Yuxiao and Wang, Kuansan and Sun, Yizhou},
    title = {Heterogeneous Graph Transformer},
    booktitle = {Proceedings of The Web Conference 2020},
    year = {2020},
    pages = {2704–2710}
}

@inproceedings{rf50,
  author    = {Velickovic, Petar and Cucurull, Guillem and Casanova, Arantxa and Romero, Adriana and Lio, Pietro and Bengio, Yoshua},
  title     = {Graph Attention Networks},
  booktitle = {Proceedings of the 6th International Conference on Learning Representations (ICLR)},
  year      = {2018}
}

@inproceedings{rf51,
  author    = {Zhu, Yun and Wang, Yaoke and Shi, Haizhou and Tang, Siliang},
  title     = {Efficient Tuning and Inference for Large Language Models on Textual Graphs},
  booktitle = {Proceedings of the 33rd International Joint Conference on Artificial Intelligence (IJCAI)},
  year      = {2024}
}

@inproceedings{rf52,
  author    = {Wu, Xixi and Shen, Yifei and Ge, Fangzhou and Shan, Caihua and Jiao, Yizhu and Sun, Xiangguo and Cheng, Hong},
  title     = {When Do LLMs Help With Node Classification? A Comprehensive Analysis},
  booktitle = {Proceedings of the 42nd International Conference on Machine Learning (ICML)},
  year      = {2025}
}

@inproceedings{rf53,
  author    = {Rossi, Emanuele and Charpentier, Bertrand and Di Giovanni, Francesco and Frasca, Fabrizio and Günnemann, Stephan and Bronstein, Michael M.},
  title     = {Edge Directionality Improves Learning on Heterophilic Graphs},
  booktitle = {Proceedings of the 3rd Annual Learning on Graphs Conference (LoG 2024)},
  volume    = {214},
  year      = {2024},
  pages     = {25:1--25:14},
}

@article{EPC,
author = {Aman Ullah and Salah Ud Din and Nasrullah Khan and Cobbinah B. Mawuli and Junming Shao},
title = {Towards investigating influencers in complex social networks using electric potential concept from a centrality perspective},
journal = {Information Fusion},
volume = {109},
pages = {102439},
year = {2024},
issn = {1566-2535}
}

@article{IDME,
author = {Zejun Sun and Yanan Sun and Xinfeng Chang and Feifei Wang and Qiming Wang and Aman Ullah and Junming Shao},
title = {Finding critical nodes in a complex network from information diffusion and Matthew effect aggregation},
journal = {Expert Systems with Applications},
volume = {233},
pages = {120927},
year = {2023},
issn = {0957-4174}
}

@article{betweenness,
  author = {Freeman, Linton C.},
  title = {A set of measures of centrality based on betweenness},
  journal = {Sociometry},
  year = {1977},
  volume = {40},
  number = {1},
  pages = {35--41}
}

@inproceedings{transformer,
 author = {Vaswani, Ashish and Shazeer, Noam and Parmar, Niki and Uszkoreit, Jakob and Jones, Llion and Gomez, Aidan N and Kaiser, \L ukasz and Polosukhin, Illia},
 title = {Attention is All you Need},
 booktitle = {Advances in Neural Information Processing Systems},
 editor = {I. Guyon and U. Von Luxburg and S. Bengio and H. Wallach and R. Fergus and S. Vishwanathan and R. Garnett},
 pages = {},
 publisher = {Curran Associates, Inc.},
 volume = {30},
 year = {2017}
}

@article{gru,
  author={Cho, Kyunghyun and Van Merri{\"e}nboer, Bart and Gulcehre, Caglar and Bahdanau, Dzmitry and Bougares, Fethi and Schwenk, Holger and Bengio, Yoshua},
  title={Learning phrase representations using RNN encoder-decoder for statistical machine translation},
  journal={arXiv preprint arXiv:1406.1078},
  year={2014}
}

@Inbook{lstm,
author="Graves, Alex",
title="Long Short-Term Memory",
bookTitle="Supervised Sequence Labelling with Recurrent Neural Networks",
publisher="Springer Berlin Heidelberg",
year="2012",
pages="37--45",
isbn="978-3-642-24797-2",
}

@InProceedings{AMIL,
  author =       {Ilse, Maximilian and Tomczak, Jakub and Welling, Max},
  title =    {Attention-based Deep Multiple Instance Learning},
  booktitle =    {Proceedings of the 35th International Conference on Machine Learning},
  series =   {Proceedings of Machine Learning Research},
  year =   {2018},
  volume =   {80},
  pages =    {2127--2136}
}

@article{od1,
author = {Xiaogang Guo and Mengyuan Fang and Luliang Tang and Zihan Kan and Xue Yang and Tao Pei and Qingquan Li and Chaokui Li},
title = {An adaptive OD flow clustering method to identify heterogeneous urban mobility trends},
journal = {Journal of Transport Geography},
year = {2025},
volume = {123},
pages = {104080},
issn = {0966-6923}
}

@article{od2,
author = {Xue Xing and Bing Wang and Xin Ning and Gang Wang and Prayag Tiwari},
title = {Short-term OD flow prediction for urban rail transit control: A multi-graph spatiotemporal fusion approach},
journal = {Information Fusion},
year = {2025},
volume = {118},
pages = {102950},
issn = {1566-2535}
}

@ARTICLE{od3,
  author={Wang, Ming and Zhang, Yong and Zhao, Xia and Hu, Yongli and Yin, Baocai},
  title={Traffic Origin-Destination Demand Prediction via Multichannel Hypergraph Convolutional Networks},
  journal={IEEE Transactions on Computational Social Systems},  
  year={2024},
  volume={11},
  number={4},
  pages={5496-5509}
}

@article{gatv2,
  title={How attentive are graph attention networks?},
  author={Brody, Shaked and Alon, Uri and Yahav, Eran},
  journal={arXiv preprint arXiv:2105.14491},
  year={2021}
}

@article{PDFormer, 
title={PDFormer: Propagation Delay-Aware Dynamic Long-Range Transformer for Traffic Flow Prediction}, 
volume={37}, 
number={4}, 
journal={Proceedings of the AAAI Conference on Artificial Intelligence}, 
author={Jiang, Jiawei and Han, Chengkai and Zhao, Wayne Xin and Wang, Jingyuan}, year={2023}, 
month={Jun.}, 
pages={4365-4373} 
}

@Article{DR-HGN,
AUTHOR = {Liu, Xiangting and Qian, Chengyuan and Zhao, Xueyang},
TITLE = {A Dynamic Regional-Aggregation-Based Heterogeneous Graph Neural Network for Traffic Prediction},
JOURNAL = {Mathematics},
VOLUME = {13},
YEAR = {2025},
NUMBER = {9},
ARTICLE-NUMBER = {1458},
ISSN = {2227-7390}
}

@article{SeHGNN, 
title={Simple and Efficient Heterogeneous Graph Neural Network}, 
volume={37}, 
number={9}, 
journal={Proceedings of the AAAI Conference on Artificial Intelligence}, 
author={Yang, Xiaocheng and Yan, Mingyu and Pan, Shirui and Ye, Xiaochun and Fan, Dongrui}, 
year={2023}, 
month={Jun.}, 
pages={10816-10824} 
}

@article{Edge-GNN,
  title={Edge-based graph neural network for ranking critical road segments in a network},
  author={Jana, Debasish and Malama, Sven and Narasimhan, Sriram and Taciroglu, Ertugrul},
  journal={Plos one},
  volume={18},
  number={12},
  pages={e0296045},
  year={2023}
}

@inproceedings{LambdaLoss,
author = {Wang, Xuanhui and Li, Cheng and Golbandi, Nadav and Bendersky, Michael and Najork, Marc},
title = {The LambdaLoss Framework for Ranking Metric Optimization},
isbn = {9781450360142},
publisher = {Association for Computing Machinery},
booktitle = {Proceedings of the 27th ACM International Conference on Information and Knowledge Management},
numpages = {10},
pages = {1313–1322},
year = {2018},
series = {CIKM '18}
}

@inproceedings{lighrgbm,
 author = {Ke, Guolin and Meng, Qi and Finley, Thomas and Wang, Taifeng and Chen, Wei and Ma, Weidong and Ye, Qiwei and Liu, Tie-Yan},
 title = {LightGBM: A Highly Efficient Gradient Boosting Decision Tree},
 booktitle = {Advances in Neural Information Processing Systems},
 editor = {I. Guyon and U. Von Luxburg and S. Bengio and H. Wallach and R. Fergus and S. Vishwanathan and R. Garnett},
 publisher = {Curran Associates, Inc.},
 volume = {30},
 year = {2017}
}

@inproceedings{Rankformer,
author = {Buyl, Maarten and Missault, Paul and Sondag, Pierre-Antoine},
title = {RankFormer: Listwise Learning-to-Rank Using Listwide Labels},
isbn = {9798400701030},
publisher = {Association for Computing Machinery},
booktitle = {Proceedings of the 29th ACM SIGKDD Conference on Knowledge Discovery and Data Mining},
numpages = {12},
pages = {3762–3773},
year = {2023},
series = {KDD '23}
}

@article{tau,
    author = {KENDALL, M. G.},
    title = {A NEW MEASURE OF RANK CORRELATION},
    journal = {Biometrika},
    volume = {30},
    number = {1-2},
    pages = {81-93},
    year = {1938},
    month = {06},
    issn = {0006-3444}
}

@inproceedings{listnet,
author = {Cao, Zhe and Qin, Tao and Liu, Tie-Yan and Tsai, Ming-Feng and Li, Hang},
title = {Learning to rank: from pairwise approach to listwise approach},
booktitle = {Proceedings of the 24th International Conference on Machine Learning},
publisher = {Association for Computing Machinery},
numpages = {8},
pages = {129–136},
year = {2007},
isbn = {9781595937933}
}

@inproceedings{mle,
  author    = {Fen Xia and Tie{-}Yan Liu and Jue Wang and Wensheng Zhang and Hang Li},
  title     = {Listwise Approach to Learning to Rank: Theory and Algorithm},
  booktitle = {Proceedings of the 25th International Conference on Machine Learning},
  series    = {ICML '08},
  year      = {2008},
  month     = jul,
  pages     = {1192--1199}
}

@article{deepsets,
  author={Zaheer, Manzil and Kottur, Satwik and Ravanbakhsh, Siamak and Poczos, Barnabas and Salakhutdinov, Russ R and Smola, Alexander J},
  title={Deep sets},
  journal={Advances in neural information processing systems},
  volume={30},
  year={2017}
}

@InProceedings{SetTrans,
  author =       {Lee, Juho and Lee, Yoonho and Kim, Jungtaek and Kosiorek, Adam and Choi, Seungjin and Teh, Yee Whye},
  title = 	 {Set Transformer: A Framework for Attention-based Permutation-Invariant Neural Networks},
  booktitle = 	 {Proceedings of the 36th International Conference on Machine Learning},
  series = 	 {Proceedings of Machine Learning Research},
  volume = 	 {97},
  pages = 	 {3744--3753},
  year = 	 {2019},
  month = 	 {09--15 Jun}
}

\end{document}